\UseRawInputEncoding
\documentclass[doublecolumn]{IEEEtran}
\usepackage{graphicx}
\usepackage{amsmath}
\usepackage{newfloat}
\usepackage{listings}
\usepackage{bbm}
\usepackage{amsthm} % definte the proof environment: block letters + italic letters
\usepackage{amssymb}
\usepackage[font=footnotesize]{caption} % set the captain font size to 8 (i.e. footnotesize)
\usepackage[noadjust]{cite} % alphabetize grouped citations automatically [3, 1, 2] to [1, 2, 3]
\usepackage{color}
\usepackage{algorithm} % options: boxed [section]
\usepackage{algorithmic}
\usepackage{enumerate}
\usepackage[hidelinks,colorlinks=false]{hyperref}
\usepackage[left=0.75in, right=0.75in, top=0.75in, bottom=0.75in]{geometry}
\usepackage{authblk} % for authors' affiliation \affil{xxx}
\usepackage{arydshln} % for dashline in table or matrix
\usepackage{multirow}
\usepackage{subfigure}
\usepackage{bm}
\usepackage{tikz}
\usepackage{pifont}% http://ctan.org/pkg/pifont
\newcommand{\cmark}{\ding{51}}%
\newcommand{\xmark}{\ding{55}}%
\usetikzlibrary{calc} % for calculation functions in Tikz let, in commands in Tikz
\usetikzlibrary{shapes} % for block diagram
\usetikzlibrary{chains}
\usetikzlibrary{fit}
\usetikzlibrary{arrows}
\usetikzlibrary{decorations.text} % text along path
\usetikzlibrary{decorations.markings}
\usetikzlibrary{decorations.pathmorphing} % for zigzag arrow
\usetikzlibrary{shadows}
\usetikzlibrary{patterns}
\usetikzlibrary{matrix}
\usepackage{pgfplots}
\usepackage[europeanresistors]{circuitikz}
\usepackage[outline]{contour} % white surround text
\contourlength{1.5pt}

\graphicspath{{figures/}}

\pdfoutput=1

\def\ie{{i.e.}}
\def\eg{{e.g.}}

\begin{document}

\title{Domain Adaptive Detection of MAVs: A Benchmark and Noise Suppression Network}
\author{Yin Zhang, Jinhong Deng, Peidong Liu, Wen Li, Shiyu Zhao
\thanks{This work was supported by the Hangzhou Key Technology Research and Development Program (Grant No. 20212013B09) and the Research Center for Industries of the Future at Westlake University (Grant No. WU2022C027). 
 (Corresponding author: Shiyu Zhao.)}
\thanks{Yin Zhang is with the College of Computer Science and Technology, Zhejiang University, Hangzhou 310027, China, and also with the School of Engineering, Westlake University, Hangzhou 310024, China (e-mail: zhangyin@westlake.edu.cn).}
\thanks{Jinhong Deng and Wen Li are with the School of Computer Science and Engineering, University of Electronic Science and Technology of China, Chengdu 611731, China (e-mail: jhdeng1997@gmail.com; liwen@uestc.edu.cn).}
\thanks{Peidong Liu is with the School of Engineering, Westlake University, Hangzhou 310024, China (e-mail: liupeidong@westlake.edu.cn).}
\thanks{Shiyu Zhao is with the Research Center for Industries of the Future and School of Engineering at Westlake University and Westlake Institute for Advanced Study, Hangzhou 310024, China (e-mail: zhaoshiyu@westlake.edu.cn).}
}

\maketitle
\begin{abstract}
Visual detection of Micro Air Vehicles (MAVs) has attracted increasing attention in recent years due to its important application in various tasks.
The existing methods for MAV detection assume that the training set and testing set have the same distribution. As a result, when deployed in new domains, the detectors would have a significant performance degradation due to domain discrepancy. In this paper, we study the problem of cross-domain MAV detection. The contributions of this paper are threefold.
1) We propose a Multi-MAV-Multi-Domain (M3D) dataset consisting of both simulation and realistic images. Compared to other existing datasets, the proposed one is more comprehensive in the sense that it covers rich scenes, diverse MAV types, and various viewing angles. A new benchmark for cross-domain MAV detection is proposed based on the proposed dataset.
2) We propose a Noise Suppression Network (NSN) based on the framework of pseudo-labeling and a large-to-small training procedure. 
To reduce the challenging pseudo-label noises, two novel modules are designed in this network.
The first is a prior-based curriculum learning module for allocating adaptive thresholds for pseudo labels with different difficulties. The second is a masked copy-paste augmentation module for pasting truly-labeled MAVs on unlabeled target images and thus decreasing pseudo-label noises.
3) Extensive experimental results verify the superior performance of the proposed method compared to the state-of-the-art ones. In particular, it achieves mAP of 46.9\%(+5.8\%), 50.5\%(+3.7\%), and 61.5\%(+11.3\%) on the tasks of simulation-to-real adaptation, cross-scene adaptation, and cross-camera adaptation, respectively. 
\end{abstract}
\def\abstractname{Note to Practitioners}
\begin{abstract}
To study the cross-domain MAV detection problem, this paper establishes a novel benchmark that consists of three domain adaptation tasks: simulation-to-real adaptation, cross-scene adaptation, and cross-camera adaptation, respectively. The benchmark is based on a novel MAV dataset called Multi-MAV-Multi-Domain (M3D), which is available at: https://github.com/WestlakeAerialRobotics/M3D. To reduce the noises caused by pseudo labels, a noise suppression network is proposed to overcome the error accumulation. Extensive experiments are conducted to prove the effectiveness of the proposed approach.
\end{abstract}

\section{Introduction}

MAV detection \cite{li2021fast, wang2021deep, isaac2021unmanned} has attracted increasing attention due to its important applications in various tasks such as multi-MAV swarming \cite{pavliv2021tracking} and detection of malicious MAVs \cite{zheng2022detection}. Although in recent years vision-based MAV detection has been studied under different setups \cite{jamil2020malicious,chen2017deep,zhao2022vision,sun2020tib}, MAV detection still faces critical challenges to be applied in practice \cite{zheng2021air}.
The existing MAV detection methods aim to improve the performance of MAV detection under the assumption that the training set and testing set have the same distribution. As a result, deploying these detectors in new domains (e.g., different environments) would result in significant performance degradation due to the domain discrepancy. To this end, it is necessary to develop novel cross-domain MAV detection approaches.

Cross-domain object detection has been studied in recent years. The benchmarks in this area contain inverse weather adaptation, simulation-to-realistic adaptation, real-to-artistic adaptation, and day-to-night adaptation \cite{tian2021knowledge, zhao2022task, ye2022unsupervised}. 
Due to the small amount of data in some target domains, a few works even exceed the supervised learning approaches \cite{li2022cross}. 
Existing methods for cross-domain object detection can be classified into domain mapping, adversarial learning, consistency learning, and pseudo-labeling. For the works focusing on the style differences, domain-mapping methods (\eg, image-to-image translation) are adopted in the adaptation algorithms to reduce the appearance gap between the source domain and target domain \cite{deng2021unbiased, zhu2017unpaired}. Adversarial learning has been widely explored to minimize domain discrepancy \cite{chen2018domain, li2022cross}. Some works propose consistency-based methods to further improve the adaptation performance \cite{deng2023harmonious}. Pseudo-labeling has become popular due to its simplicity and effectiveness despite the performance being restricted by the pseudo-label noises \cite{ramamonjison2021simrod}.

\begin{figure*}[t]
	\centering
	\includegraphics[width=1\textwidth]{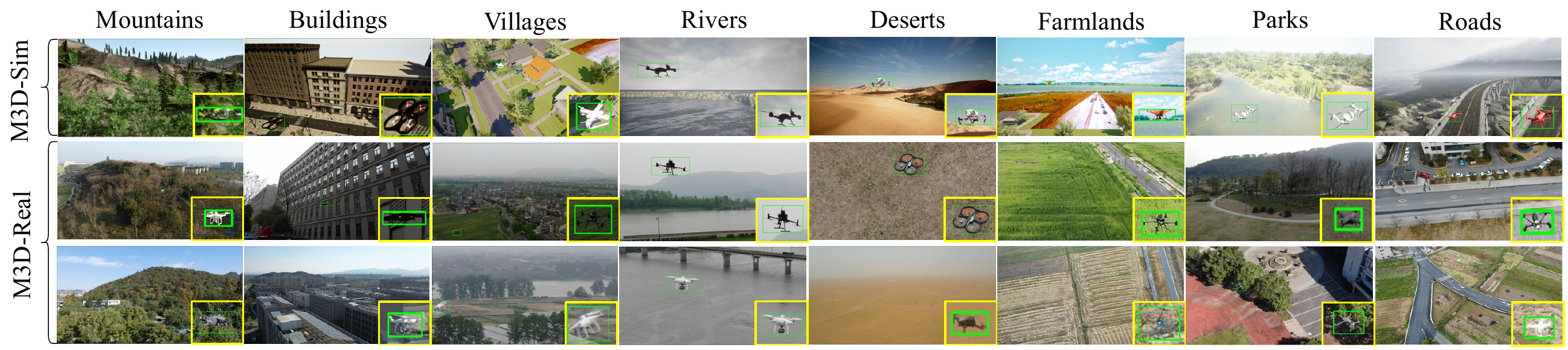}
	\caption{Samples from the proposed Multi-MAV-Multi-Domain (M3D) dataset. The top to bottom shows examples from the M3D-Sim subset and M3D-Real subset, respectively. The backgrounds of MAVs contain many types: mountains, buildings, villages, rivers, deserts, farmlands, parks, and roads. MAVs are usually small objects in captured images. The enlarged MAVs containing diverse types are presented at the lower right of each picture.}
	\label{Sim-RealDrone}
\end{figure*}

While advancements have been made in cross-domain object detection, it is crucial to notice that the challenges specific to MAV detection require tailored solutions. 
First, the backgrounds of MAVs are more diverse than common objects like cars or pedestrians. As shown in Fig.~\ref{Sim-RealDrone}, MAVs captured by cameras with multi-viewing angles (\ie, ground-to-air and air-to-air) may appear in diverse scenes. The diversity and complexity of the backgrounds cause severe challenges for MAV detection.
Second, due to the requirement for long-range detection, the imaging sizes of MAVs are usually much smaller than that of the objects in natural images. For example, the average area of the objects in the real-to-artistic adaptation task occupies 12.7\% of the entire image \cite{inoue2018cross}, while the MAVs in Drone-vs-Bird dataset only take 0.10\% \cite{coluccia2021drone}. Small objects further increase the difficulty of detection.
Finally, MAV detection requires real-time detection, which is an issue that is usually ignored by cross-domain object detection researchers cause most methods are based on two-stage detection networks with high accuracy but low efficiency. Therefore, it is necessary to develop novel cross-domain MAV detection approaches tailored to the unique characteristics and challenges of MAV detection.

In light of these challenges, this paper focuses on the cross-domain MAV detection problem. We aim to use the unlabeled target data to help the MAV detectors adapt to the target domain. The contributions of this paper are threefold.

1) We propose a Multi-MAV-Multi-Domain~(M3D) dataset to study the cross-domain MAV detection problem. Compared to the state-of-the-art MAV datasets \cite{zheng2021air, coluccia2021drone, jiang2021anti}, our dataset contains different styles of images, more diverse environments, and more types of MAVs. Our dataset consists of two subsets: M3D-Sim and M3D-Real with 28,740 and 55,259 images, respectively. Some image examples are shown in Fig.~\ref{Sim-RealDrone}. M3D-Sim is collected from a simulation platform called Unreal Engine \cite{sanders2016introduction} and includes 22 MAVs and 16 environments. M3D-Real is a realistic dataset that contains diverse scenes and 10 types of MAVs. 

Based on the proposed M3D and other existing MAV datasets, we construct a novel benchmark that contains several representative tasks: \emph{simulation-to-real adaptation}, \emph{cross-scene adaptation}, and \emph{cross-camera adaptation}, respectively. A comprehensive experiment is also conducted on the M3D dataset for universal MAV detection. To our best knowledge, we are among the first to benchmark the problem of cross-domain MAV detection and study this issue in this area.

2) We propose a novel Noise Suppression Network (NSN) to address the domain discrepancy through pseudo-labeling. The inaccuracy of the pseudo labels will increase when meeting smaller MAVs under more diverse backgrounds. NSN can effectively overcome the limitations of existing unsupervised domain adaptation methods. NSN consists of two modules, a prior-guided curriculum learning module, and a masked copy-paste augmentation module, to fully utilize the domain-specific information of target data and reduce the pseudo-label noises.
First, the prior-guided curriculum learning module is for generating more accurate pseudo labels. A novel difficulty matrix is adopted to evaluate the complexity degree of MAV samples.
Then, curriculum learning is used to help the model adapt to target samples with different detection difficulties.
Second, the masked copy-paste augmentation module is for decreasing the pseudo-label noises by adding more true labels. Cropped MAV images with true labels are segmented automatically first and then pasted on the unlabeled target images. The participation of true labels can help the detection model adapt to new backgrounds and reduce the influence of noises.

These two simple yet effective modules eliminate the error accumulation from self-training caused by pseudo-label noise, which impedes the performance improvements of MAV detectors on the target domain. Besides, to meet the requirement of real-time MAV detection, we also employ a large-to-small model training procedure to help the large model gradually transfer knowledge to the small model. Our NSN method is model-agnostic and can be combined with other detection networks.

3) We conduct extensive experiments on the proposed domain adaptive MAV detection benchmark. Before adaptation, the detection models face significant performance degradation and only achieve mAPs of 39.9\%, 28.7\%, and 31.9\% on the target datasets. The experimental results show the effectiveness of our proposed method. In particular, our method achieves mAPs of 46.9\%(+5.8\%), 50.5\%(+3.7\%), and 61.5\%(+11.3\%) on the tasks of simulation-to-real adaptation, cross-scene adaptation, and cross-camera adaptation, respectively, which significantly outperforms the baseline algorithm \cite{ramamonjison2021simrod}. Our method also achieves state-of-the-art results on the proposed adaptation tasks.

We also conduct ablation studies to show the functionalities of different modules in NSN. 
The ablation study contains three parts, the participation of different modules, the choice of cropped images for data augmentation, and the varying of different hyper-parameters.
We also compare our augmentation module with other methods to show the improvement ability on the domain adaptation tasks.
The qualitative and experimental analysis provides ample evidence for the feasibility and effectiveness of the proposed method.

\section{Related Work}
\subsection{ Object Detection and MAV Detection}
Object detection is a well-studied area and can be separated into one-stage-based and two-stage-based methods \cite{lu2020mimicdet}. The main difference between these two types of methods is whether the network contains a region proposal network (RPN) module \cite{ren2015faster}. Generally, one-stage-based methods cost less time but have lower accuracy. YOLO \cite{redmon2018yolov3, bochkovskiy2020yolov4}, RetinaNet \cite{lin2017focal}, FCOS \cite{tian2019fcos} and Centernet \cite{duan2019centernet} are some representative detection models. Two-stage methods like R-FCN \cite{dai2016r}, Cascade-RCNN \cite{cai2018cascade}, and DETR \cite{carion2020end} can produce results with higher accuracy but cost more time. Thus, some researchers aim to make the one-stage-based detection models inference fast with higher precision. One-stage-based methods are suitable for MAV detection because the detection models require to be deployed to embedded devices with limited computation ability.

MAV detection methods can be classified into two types: appearance-based methods \cite{vrba2020marker, pavliv2021tracking} and motion-based methods \cite{rozantsev2016detecting, jiang2021anti}. In this paper, we mainly focus on improving the adaptation performance of appearance-based methods. Many MAV detectors draw lessons from the object detection area and are built on detection models like Faster-RCNN \cite{ren2015faster} and YOLO \cite{bochkovskiy2020yolov4}. Due to the importance of MAV detection, there have been some works that designed deep-learning networks that specialized in detecting MAVs to solve this issue \cite{ashraf2021dogfight, li2023global}. \cite{nguyen2019mavnet} proposes MAVNet to achieve real-time detection of MAVs from a semantic segmentation network with fewer parameters. TIB-Net \cite{sun2020tib} utilizes a tiny iterative backbone to enhance the performance of small objects. FastUAVNet \cite{yavariabdi2021fastuav} is modified on YOLOv3 \cite{redmon2018yolov3} by using the inception blocks to extract local and glocal features. However, the performance of those MAV detectors degrades sharply when the training and testing domains are not in the same distribution. To improve the detection performance of MAV detectors in testing environments, we propose the cross-domain MAV detection benchmark for further research.

\subsection{Cross-Domain Object Detection}
Cross-domain object detection methods aim to adapt to the unlabelled target domain from the labeled source domain. The benchmarks for cross-domain detection have several types, which are concluded in Table ~\ref{TraditionalTask}. Some target datasets in these benchmarks have a small amount of data, making the domain adaptation methods exceed the supervised learning with the help of a large amount of labeled source data \cite{deng2021unbiased, li2022cross}. There are also domain adaptation works that specialize in computer vision applications like pedestrian detection \cite{jiao2021san}, face recognition \cite{faraki2021cross}, and X-ray object detection \cite{tao2022exploring}. Many methods have been proposed to address the domain discrepancy in object detection. Most cross-domain object detection methods adopt Faster-RCNN as their detection model. There are a few works that pay attention to one-stage models to study the domain adaptation issue \cite{mattolin2023confmix, ramamonjison2021simrod, deng2023harmonious, li2022scan, li2022sigma}.

\begin{table}[htbp] \centering
	\renewcommand{\arraystretch}{1.15}
	\begin{tabular}{cccc}
		\hline \hline
		\textbf{Scenarios} & \textbf{Datasets} & \textbf{$N_{S}$} & \textbf{$N_{T}$} \\\hline
		Simulation-to-real  & SIM10K vs Cityscapes  & 10,000 & 5,000 \\
		Normal-to-foggy  &  Cityscapes vs Foggyscapes & 5,000 & 5,000 \\
		Different cameras & KITTI vs Cityscapes & 7,481 & 5,000\\
		Real-to-artistic &  PASCAL VOC vs Clipart & 21,503 & 1,000 \\
		\hline \hline
	\end{tabular}
	\caption{The comparison of different tasks in traditional domain adaptation benchmark. $N_{S}$ and $N_{T}$ represent the image numbers in the source dataset and target dataset.}
	\label{TraditionalTask}
\end{table}

The cross-domain object detection methods \cite{oza2023unsupervised} could be mainly divided into adversarial training \cite{chen2018domain, saito2019strong, li2022cross}, domain alignment \cite{cai2019exploring, saito2019strong, wang2021afan, zhao2022task}, and self-training \cite{roychowdhury2019automatic, khodabandeh2019robust}. Adversarial learning-based methods are suitable to deal with domain adaptation tasks with huge differences.  The NLTE method \cite{liu2022towards} proposes a data mining module and a graph relation module for learning the domain-invariant representations. When the target data doesn't have significant differences from the source data, adversarial training may be ineffective in learning domain-invariant representations. Domain alignment-based methods aim to reduce the distribution discrepancy by aligning features at the region level or instance level. The AsyFOD method \cite{gao2023asyfod} aims to reduce the domain gap by conducting the feature alignment between target-dissimilar source instances and augmented
target instances to avoid over-adaptation. The MGADA method \cite{zhou2022multi} tries to align the source and target domains from the pixel level, instance level, and category level. However, these two types of methods do not fully exploit the domain-specific information of the target data.  

Data augmentation algorithms are widely utilized in cross-domain detection and have many types: GAN-based methods, strong-weak augmentation and mixup augmentation. Many works \cite{kim2019diversify, gong2019dlow, huang2018auggan} use GAN-based methods (\eg, CycleGAN \cite{zhu2017unpaired}) to reduce the gap between different domains. However, this type of method can have satisfactory results in tackling style-difference domains but has poor performance in solving real-to-real or scene-to-scene issues. Strong-weak augmentation methods usually appear with the teacher-student models to learn the appearance-invariant features \cite{xu2021end}. Some works \cite{wang2021afan} use mixup augmentation \cite{zhang2018mixup} to generate domain intermediate images.  The AcroFOD method \cite{gao2022acrofod} proposes a multi-level data augmentation algorithm and a new training strategy to reduce the domain shift. The above methods all attempt to reduce the shifts between source and target domains by making their appearances look more similar. However, the unlabeled target data is still not fully utilized in an unsupervised learning way.
\begin{figure*}[t]
	\centering
	\subfigure[The 22 types of MAVs in M3D-Sim subset;]{
		\centering
		\includegraphics[width=1\textwidth]{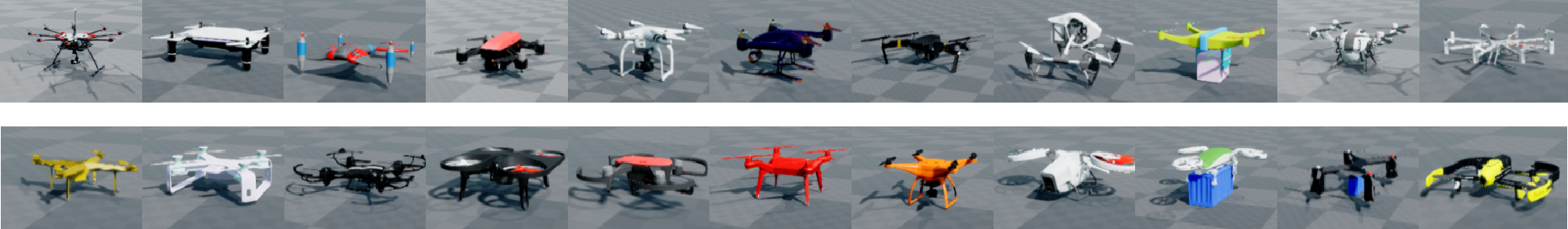}
	}
	\centering
	\subfigure[The 16 types of simulation environments in M3D-Sim subset.]{
		\centering
		\includegraphics[width=1\textwidth]{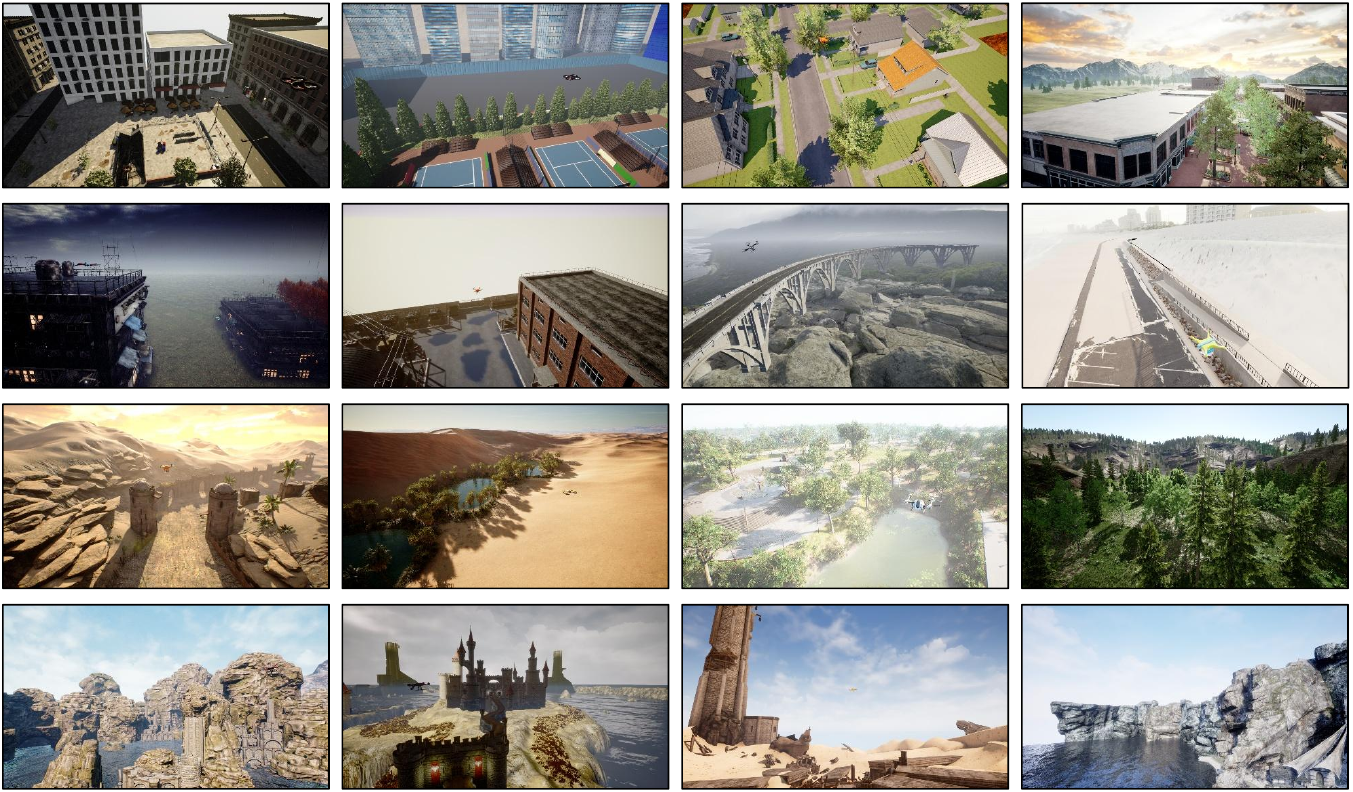}
	}
	\caption{The MAVs and environments in M3D-Sim subset.}
	\label{Airsim}
\end{figure*}

\begin{figure*}[t]
	\centering
	\subfigure[The 10 types of MAVs in M3D-Real subset;]{
		\centering
		\includegraphics[width=1\textwidth]{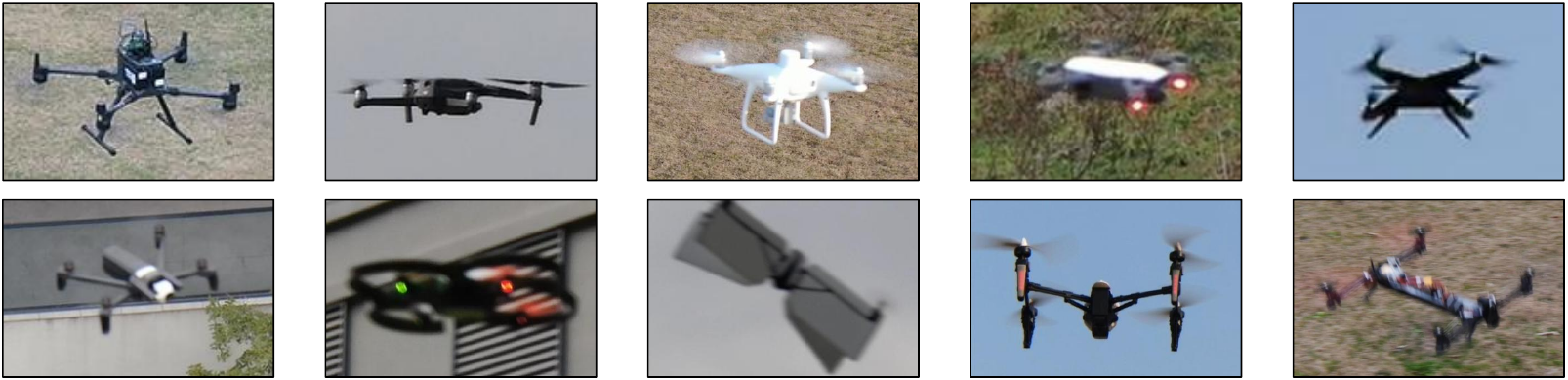}
		\label{DRones}
	}
	\centering
	\subfigure[The 20 examples of environments in M3D-Real subset.]{
		\centering
		\includegraphics[width=1\textwidth]{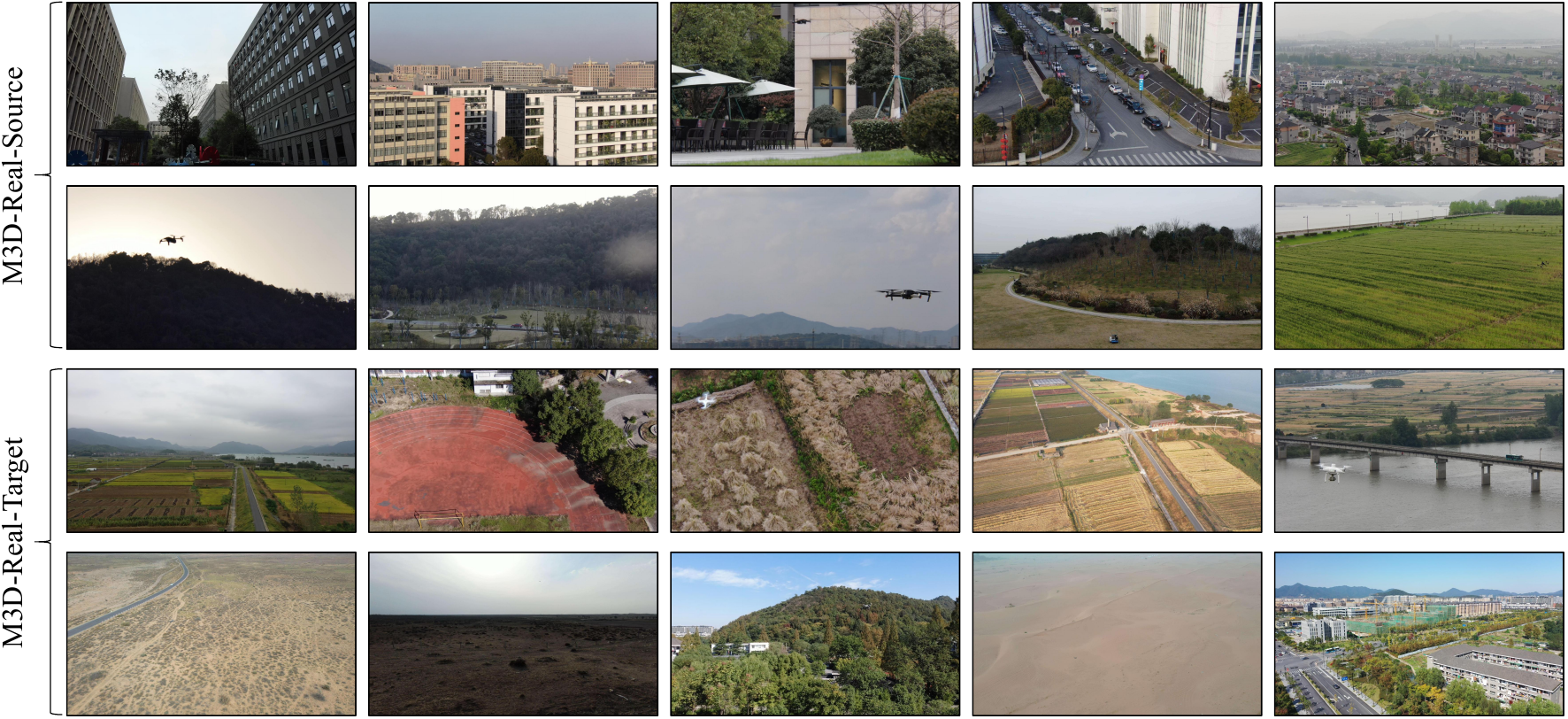}
		\label{RealScenes}
	}
	\caption{The MAVs and environments in M3D-Real subset.}
\end{figure*}

\subsection{Pseudo-Labeling in Domain Adaptation}
Pseudo-labeling is a popular way to deal with semi-supervised learning and unsupervised learning tasks due to its simplicity and effectiveness. For domain adaptation tasks, the pseudo-labeling technique has been widely combined with several types of methods to reduce the noises and obtain pseudo-labels with higher accuracy. The TDD method \cite{he2022cross} is based on a unified teacher-student learning framework and a target proposal perceiver module to reduce the domain shift. The SimROD method \cite{ramamonjison2021simrod} proposes to use the heavy-weight teacher model to generate more precise pseudo labels. However, when there exists a large domain shift, the teacher model still cannot generate reliable pseudo labels for training. Some works utilize curriculum learning to improve the accuracy of pseudo labels and obtain better performances \cite{zhang2017curriculum, soviany2021curriculum}. Compared to traditional deep learning, curriculum learning is a novel training paradigm by learning from easy to hard examples \cite{soviany2022curriculum}. However, due to the particularity of the MAV detection problem, existing curriculum domain adaptation methods are less effective in reducing the domain gap. Some works refine pseudo labels by exploiting the intra-class similarity and aligning the source and target domains \cite{wang2023improving}. Due to the diversity of MAVs, backgrounds, and perspectives, the similarity of the MAV category is non-centralized and hard to utilize.

To avoid recurrent problems, we propose an MAV dataset containing a large amount of data and build a cross-domain MAV detection benchmark based on the dataset. We pay attention to the pseudo-labeling noise generated by self-training methods. The noise of pseudo labels increases when meeting small MAV targets and diverse backgrounds. Two modules are proposed to reduce the noise and help the detection model adapt to target domains better.
\section{Multi-MAV-Multi-Domain Dataset}
We propose an MAV dataset called Multi-MAV-Multi-Domain (M3D) to study the domain adaptation problem. The M3D dataset contains two types of images: simulation images (M3D-Sim) and realistic images (M3D-Real). This dataset contains diverse MAVs, scenes, and viewing angles in both the simulation set and the real-world set. Some examples of these two sets are shown in Fig.~\ref{Sim-RealDrone}. Our dataset only contains the MAV class. We do not label other objects so there is only the MAV class in the images. There are a total of 83,999 images with 86,415 boxes in the M3D dataset. Thus there is one MAV average per image. The data is collected under the condition that one MAV pursues another MAV in the air. The image numbers of these subsets are shown in Table~\ref{Subsets}. Details of the subsets are introduced as follows.

\begin{table}[htbp] \centering
	\renewcommand{\arraystretch}{1.15}
	\begin{tabular}{cccc}
		\hline \hline
		\textbf{Subsets} & \textbf{Training Set} & \textbf{Validation Set} & \textbf{Testing Set} \\\hline
		M3D-Sim  & 22,992 & 5,748 & -\\
		M3D-Real  & 35,081 & 5,018 & 15,215\\
		\hline
		M3D-Sim-Normal &  14,855 & 3,621 & -\\
		M3D-Sim-Texture & 8,137  & 2,127 & -\\
		\hline
		M3D-Real-Source &  24,441 & 3,520 & 12,027\\
		M3D-Real-Target & 10,640 & 1,498 & 3,188\\
		\hline \hline
	\end{tabular}
	\caption{The image numbers of different subsets in our M3D dataset.}
	\label{Subsets}
\end{table}

\begin{table*}[htbp] \centering
	\renewcommand{\arraystretch}{1.15}
	\begin{tabular}{lrrrrrrr}
		\hline \hline
		\textbf{Attributes} & \textbf{Det-Fly}\cite{zheng2021air} & \textbf{Real-World}\cite{pawelczyk2020real} & \textbf{Drone-vs-Bird} \cite{coluccia2021drone} & \textbf{MIDGARD} \cite{walter2020training} & \textbf{USC-Drone} \cite{chen2017deep} & \textbf{YouTube} \cite{chen2017deep} & \textbf{Ours}\\
		\hline
		Images    & 13,271 & 55,446 & 19,027 & 8,775  & 18,778 & 1,764 & 83,999 \\
		Boxes     & 13,271 & 55,539 & 23,335  & 10,696 & 1,264  & 1,675 & 86,415 \\
		Data Source & Institute & Internet & Institute & Institute & Institute & YouTube & Institute\\
		MAV Types & 1 & - & 7 & 1 & 1 & - & 22+10 \\
		Simulation & \xmark  &\xmark&\xmark&\xmark&\xmark  &\xmark&\cmark\\
		Air-to-Air&      \cmark  &\xmark&\xmark&\cmark&\xmark  &\xmark&\cmark\\
		Ground-to-Air&   \xmark  &\cmark&\cmark&\cmark&\cmark  &\cmark&\cmark\\
		Air-to-Ground&   \cmark  &\xmark&\xmark&\xmark&\xmark  &\xmark&\cmark\\
		Moving Camera &  \cmark  &\cmark&\xmark&\cmark&\cmark  &\cmark&\cmark\\
		\hline \hline
	\end{tabular}
	\caption{The comparison of different MAV datasets. ``Air-to-air" and ``Air-to-ground" represent that the images are captured by a camera fixed on an MAV in the air.}
	\label{Comparison}
\end{table*}

\subsection{Simulation Subset: M3D-Sim}
M3D-Sim subset contains 28,740 simulation images. There are 22,992 images in the training set and 5,748 images in the validation set, respectively. To reduce the time and cost caused by human labeling, we use the simulation platform to automatically collect data. Multiple MAVs and environments are included in M3D-Sim to improve the generalization ability of the simulation set. With the help of Unreal Engine and Airsim \cite{shah2018airsim}, we collect data on 22 types of MAVs and 16 different environments, whose examples are shown in Fig.~\ref{Airsim}. The images are captured from multiple perspectives (\ie, ground-to-air, air-to-air, and air-to-ground) by the same camera fixed on another MAV. The images collected in the 16 environments establish a subset called M3D-Sim-Normal. Besides, the partial dataset is generated by texture randomization \cite{loquercio2019deep} to reduce the gap between simulation and reality. Therefore, the M3D-Sim dataset consists of two subsets: M3D-Sim-Normal and M3D-Sim-Texture, containing 18,476 and 10,264 images, respectively. The purpose of the M3D-Sim-Texture subset is to validate the feasibility of texture randomization methods. Details are shown in Section~\ref{sec:texture}. Compared to the state-of-the-art dataset in \cite{rui2021comprehensive}, our dataset contains more types of MAVs and more diverse environments with a moving camera and multi-viewing angles.

\subsection{Realistic Subset: M3D-Real}
The M3D-Real subset contains 55,259 realistic images. Unlike common objects like cars or pedestrians, the collection of MAV images requires the preparation of multiple types of MAVs and a moving camera (\eg, another MAV with a camera). The motion of the camera can guarantee the diversity of shooting perspectives and changes in backgrounds. Thus the data collection of MAVs is more arduous and time-consuming. In the M3D-Real subset, we purchased 10 types of MAVs, which are shown in Fig.~\ref{DRones}. These MAVs have different sizes, shapes, and poses in the images. The images are collected in two cities with different natural styles. We collect data at multiple locations whose categories contain mountains, buildings, villages, rivers, deserts, farmlands, parks, and roads. Each category also has several different locations for diversity. Some examples are shown in Fig.~\ref{RealScenes}. The trajectory of the MAVs at each location covers an area whose maximum value achieves one square kilometer. Thus the backgrounds of this real subset are much more diverse compared to existing MAV datasets collected by institutes. These characteristics can increase the generalization of the MAV detection network.

To study the impact of background changes on detection, all images are captured by the same camera (DJI H20T). Benefiting from the wide variety of environments, we separate the dataset into a source domain (M3D-Real-Source) and a target domain (M3D-Real-Target) to study the scene-adaptation problem. Fig.~\ref{RealScenes} shows some examples from the two sets for comparison.  The M3D-Real-Source subset and the M3D-Real-Target subset contain 39,933 images and 15,326 images, respectively.

\subsection{Comparison with Other MAV Datasets}
The comparison of different MAV datasets is shown in Table~\ref{Comparison}. The table contains six representing datasets. We compared various attributes of these datasets, such as the number of images, data sources, and shooting perspectives. Det-Fly dataset \cite{zheng2021air} involves multiple shooting angles but only one type of MAV. Real-World dataset \cite{pawelczyk2020real} has multiple MAVs, but all the images are collected from the Internet. Besides, the images are mainly shot from the ground-to-air perspective. Most of the images in the Drone-vs-Bird dataset \cite{coluccia2021drone} have small MAVs, but the camera is fixed on the ground. MIDGARD dataset \cite{walter2020training} contains indoor and outdoor environments with a single MAV and low-resolution images. There is only one MAV (DJI Phantom) in the USC-Drone dataset \cite{chen2017deep} with a ground-to-air viewing angle. Besides, only one-fifteenth of the data in this dataset is labeled. YouTube dataset \cite{chen2017deep} only contains a small number of images collected from YouTube.

Our M3D dataset has several advantages. First, our dataset contains images from multi-domains. Both the simulation set and realistic set contain large amounts of images. Thus it is suitable for studying cross-domain MAV detection. Second, compared to other datasets, our dataset contains more types of MAVs and more diverse environments. Third, all the images are captured by a moving camera with multiple viewing angles. The diversity of viewing angles guarantees the generality of the data. Due to the diversity of our dataset, we conduct several types of domain adaptation tasks, which are introduced in Section~\ref{tasks}.

\section{Proposed Method}

The motivation of this work is to design a cross-domain MAV detection approach that can reduce domain discrepancy so that the model can perform well in the target domain. Suppose that the source domain $\{\mathcal{S}, \mathcal{Y}_s\} = \{(I^s_i,y_i)|_{i=1}^{N_s}\}$ has $N_s$ labeled samples and the target domain $\mathcal{T}=\{I^t_i|_{i=1}^{N_t}\}$ has $N_t$ unlabeled samples, where $I_i$ is the $i$-th image and $y_i$ is the corresponding labels consisting of bounding boxes and category information. The source and the target domains have different but related distributions.
For the cross-domain MAV detection task, we aim to obtain an adaptive detection model $\mathcal{M}(\mathcal{S, T})$ by using both the labeled source domain and unlabeled target domain samples.

We use self-training strategies to address the domain discrepancy of cross-domain MAV detection. Here, pseudo-label noises are restrictions to improve the model performance. Thus, we propose a novel Noise Suppression Network (NSN) to suppress the noises from both internal and external ways. NSN is based on the framework of pseudo-labeling and a large-to-small training procedure. This network consists of two novel modules. Section ~\ref{sec:proposed_approach} presents the overall training procedure of our model. Then, we elaborate on two critical modules of our model, i.e., the Prior-guided Curriculum Leaning~(PCL) in Section~\ref{sec:pcl} and Masked Copy-paste Augmentation~(MCA) in Section~\ref{sec:mca}.

\subsection{Overall Approach}
\begin{figure*}[t]
	\centering
	\includegraphics[width=1\textwidth]{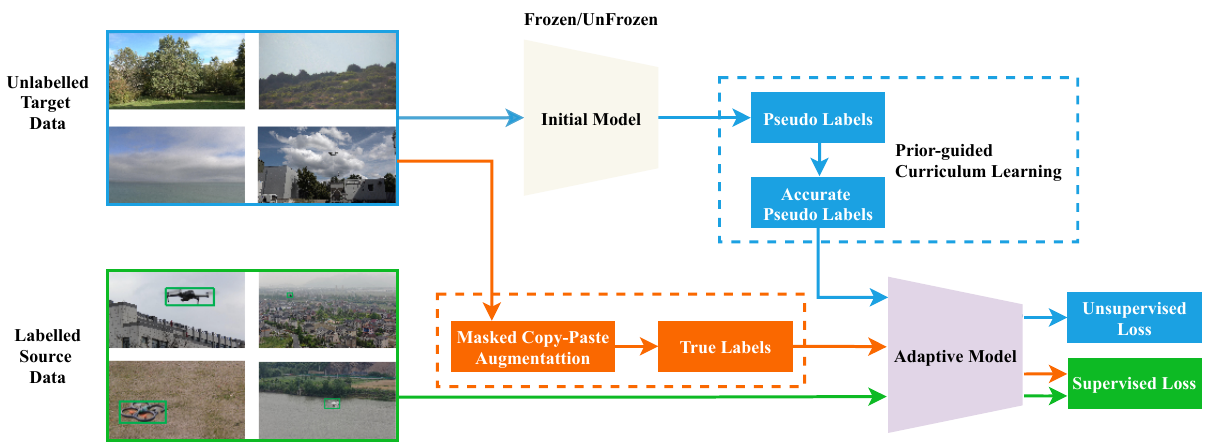}
	\caption{The framework of each training stage in the noise suppression network. 1) The prediction of unlabeled target data is employed as pseudo labels and corrected by the prior-guided curriculum learning module; (see Section~\ref{sec:pcl}) 2) The unlabeled target data is augmented by the masked copy-paste augmentation module that can generate true labels; (see Section~\ref{sec:mca}) 3) The detection model is trained by labeled source data and augmented target data for $E$ epochs.}
	\label{Drone Adaptor}
\end{figure*}

\label{sec:proposed_approach}

NSN contains a large-to-small model training procedure to transfer the knowledge of a heavyweight model to a lightweight model. Thus, we can obtain a small but accurate adaptive model capable of real-time detection. The training procedure contains three stages and is explained in Algorithm~\ref{NSN}.

\begin{algorithm}[htbp]
	\caption{Noise Suppression Network (NSN)}
	\label{NSN}
	\textbf{Input}: Labeled source dataset {$\{\mathcal{S,Y_S}\}$}, Unlabeled target images $\mathcal{T}$ \\
	\textbf{Parameter}: Training epochs $E$ \\
	\textbf{Output}: Adaptive model $\mathcal{M}_s(\mathcal{S,T})$
	
	\begin{algorithmic}[1]
		\STATE \textbf{Stage 1:} Train source model $\mathcal{M}_l(\mathcal{S})$ and $\mathcal{M}_s(\mathcal{S})$. \\
		\STATE \textbf{Stage 2.1:} Generate pseudo labels $\mathcal{Y}^{p}_\mathcal{T}$=$F_{\mathcal{M}_l}(\mathcal{T})$;\\
		\STATE Use the PCL module to correct pseudo labels $\mathcal{Y}^{p}_\mathcal{T}$;\\
		\STATE Use MCA to augment $\mathcal{T}_\mathrm{A}$ and add true labels $\mathcal{Y}^{t}_\mathcal{T}$;\\
		\STATE Freeze $w_{\mathcal{M}_l}$ except for BatchNorm Layers; \\
		\STATE Use $\{\mathcal{Y_S | S}$, ($\mathcal{Y}^{t}_\mathcal{T}$, $\mathcal{Y}^{p}_\mathcal{T}$)$|\mathcal{T}_\mathrm{A}\}$ to train $\mathcal{M}_l(\mathcal{S},\mathcal{T}_\mathrm{A})$ for $E$.\\
		\STATE \textbf{Stage 2.2:} Repeat lines 2-4 based on the last model and update $\mathcal{Y}^{t}_\mathcal{T}$, $\mathcal{Y}^{p}_\mathcal{T}$, $\mathcal{T}_\mathrm{A}$ ;
		\STATE Unfreeze the $w_{\mathcal{M}_l}$ of $\mathcal{M}_l(\mathcal{S}, \mathcal{T}_\mathrm{A})$; \\
		\STATE Use $\{\mathcal{Y_S | S}$, ($\mathcal{Y}^{t}_\mathcal{T}$, $\mathcal{Y}^{p}_\mathcal{T}$)$|\mathcal{T}_\mathrm{A}\}$ to train $\mathcal{M}_l(\mathcal{S},\mathcal{T}_\mathrm{A})$ for $E$.\\
		\STATE \textbf{Stage 3:} Repeat lines 2-4 based on the last model and update $\mathcal{Y}^{t}_\mathcal{T}$, $\mathcal{Y}^{p}_\mathcal{T}$, $\mathcal{T}_\mathrm{A}$ ;
		\STATE Freeze $w_{\mathcal{M}_s}$ except for BatchNorm Layers; \\
		\STATE Use $\{\mathcal{Y_S | S}$, ($\mathcal{Y}^{t}_\mathcal{T}$, $\mathcal{Y}^{p}_\mathcal{T}$)$|\mathcal{T}_\mathrm{A}\}$ to train $\mathcal{M}_s(\mathcal{S},\mathcal{T}_\mathrm{A})$ for $E$.\\
		\STATE \textbf{return} $\mathcal{M}_s(\mathcal{S},\mathcal{T}_\mathrm{A})$
	\end{algorithmic}
\end{algorithm}

\textbf{Stage 1:} Denote the large model and the small model as $\mathcal{M}_l$ and $\mathcal{M}_s$, respectively. We first train these two models on the source domain and obtain a source large model $\mathcal{M}_l(\mathcal{S})$ and a source small model $\mathcal{M}_s(\mathcal{S})$.

\textbf{Stage 2:} The source large model $\mathcal{M}_l(\mathcal{S})$ generates initial pseudo labels $\mathcal{Y}^{p}_\mathcal{T}$ of the unlabeled target data. To improve the accuracy of pseudo labels, we use the prior-based curriculum learning module (Section~\ref{sec:pcl}) to correct them. Moreover, the masked copy-paste module (Section~\ref{sec:mca}) is adopted to reduce the pseudo-label noises by generating augmented target images $\mathcal{T}_\mathrm{A}$. The adaptive model $\mathcal{M}_l(\mathcal{S},\mathcal{T}_\mathrm{A})$ is obtained by training the source data and target data. To make the best of the large model, we retrain the model  $\mathcal{M}_l(\mathcal{S},\mathcal{T}_\mathrm{A})$ using a new set of data with updated pseudo-labels which are more precise than the last period. Stage 2 contains two periods. The weight $w_{\mathcal{M}_l}$ of the model is frozen except for BatchNorm Layers at the first training period. The weight is unfrozen during the second training period. In this way, the adaptive large model can perform well on the target domain.

\textbf{Stage 3:} The adaptive large model generates more precise pseudo labels to train the small model. To transfer the knowledge from the large model to the small model, we still follow the same training steps to train the lightweight model. The small model $\mathcal{M}_s(\mathcal{S},\mathcal{T}_\mathrm{A})$ is eventually obtained.

The implementation procedure is shown in Algorithm~\ref{NSN}. Stage 2.1, Stage 2.2, and Stage 3 contain the same data preparation and training procedure. The training framework of these three periods is illustrated in Fig.~\ref{Drone Adaptor}. The training loss consists of a supervised loss from the labeled source data and an unsupervised loss from the unlabeled target data. The prior-guided curriculum learning module is utilized to generate more accurate pseudo labels. The masked copy-paste module provides a set of true labels to reduce the proportion of pseudo-labels in training loss. Instead of updating the pseudo-labels and augmented images at each training epoch, we choose to update data only once at the initialization of each period. The detection model can stably learn features of true labels in this way.

\subsection{Prior-Guided Curriculum Learning}
\label{sec:pcl}
In this section, we propose a Prior-guided Curriculum Learning module (PCL) to improve the accuracy of pseudo labels.  Since the prediction confidences of simple examples are much higher than that of hard examples, PCL corrects pseudo labels by assigning different difficulties of examples adaptive thresholds instead of a constant threshold. First, we use prior knowledge of MAV detection to partition pseudo labels of the target dataset into subsets of different difficulties. Second, we allocate adaptive thresholds to pseudo labels in different subsets. The details are as follows.
\subsubsection{Dataset partition}
How to evaluate the detection difficulty of targets is the key issue for curriculum learning. We propose a  \emph{difficulty matrix} to cluster different categories of a MAV dataset. Many factors influence the detection of MAVs, such as the target size, local contrast, and background complexity. The calculation of the difficulty matrix is shown in the following.

\textbf{Target size.} The target size influences the MAV detection performance especially when MAVs are extremely small objects in images. The target area is a bounding box with a height $h_\mathrm{t}$ and a width $w_\mathrm{t}$. We select the target size $m_\mathrm{ts}$ as a difficulty metric:
	\begin{equation}
		m_\mathrm{ts}=h_\mathrm{t} w_\mathrm{t}.
	\end{equation}
	
\begin{figure}[t]
	\centering
	\includegraphics[width=0.5\columnwidth]{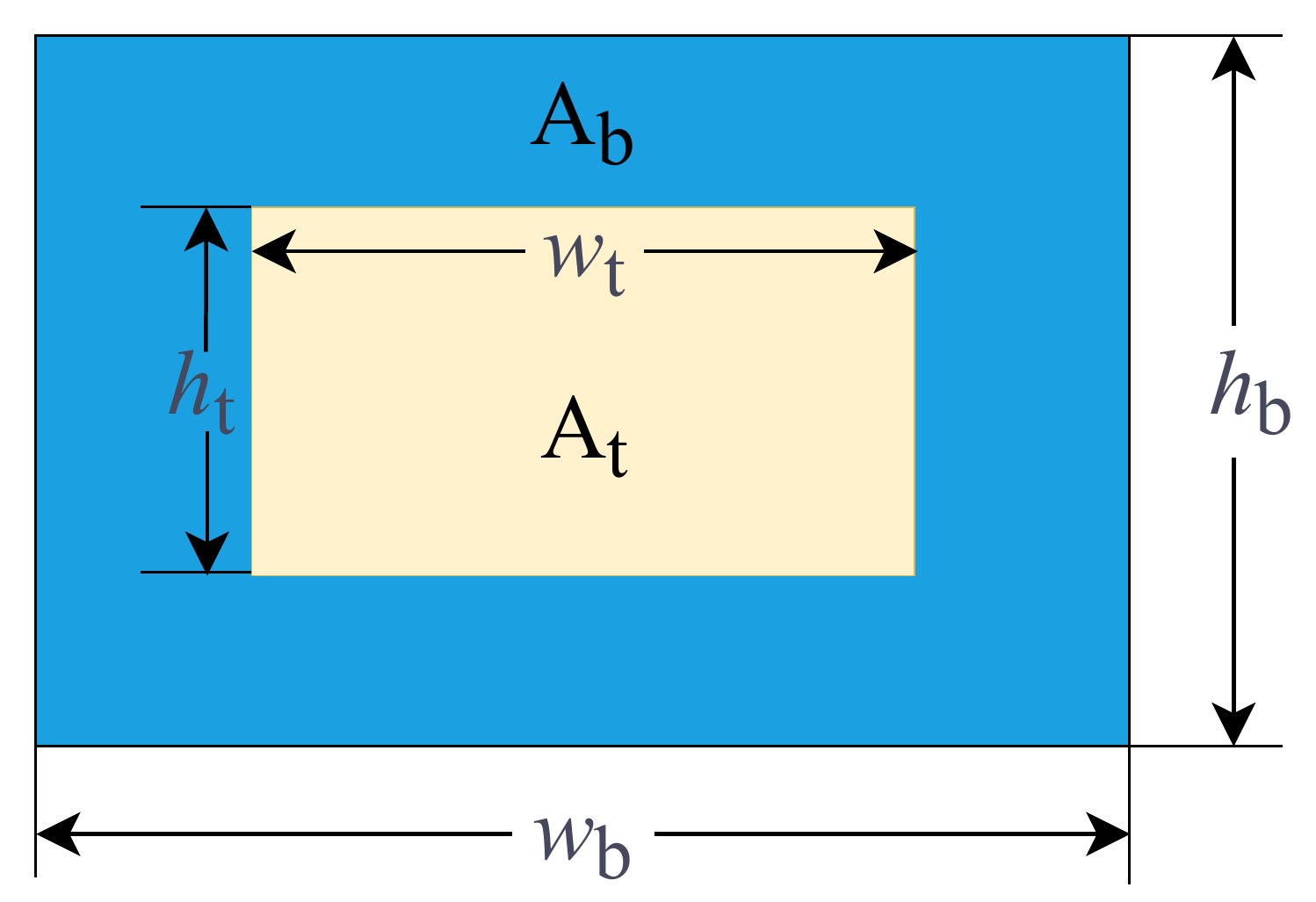}
	\caption{Schematic diagram of the target area $A_\mathrm{t}$ and local background  area $A_\mathrm{t}$. The yellow area and the blue area represent $A_\mathrm{t}$ and $A_\mathrm{b}$, respectively. }
	\label{Complexity}
\end{figure}

\textbf{Local contrast.} The contrast between the local background and the target area is also another factor that influences the detection performance. The schematic diagram of the target area and local background area is shown in Fig.~\ref{Complexity}, where $A_\mathrm{b}$ and $A_\mathrm{t}$ represent the total pixels in the local background area and target area, respectively. We define $N_\mathrm{b}$ as the number of pixels in $A_\mathrm{b}$, which can be represented as
\begin{equation}
	N_\mathrm{b} = h_\mathrm{b} w_\mathrm{b} - h_\mathrm{t} w_\mathrm{t},
\end{equation}
where $h_\mathrm{b}$ and $w_\mathrm{b}$ are the height and width of the local background area.
We select the local contrast $m_\mathrm{lc}$ between $A_\mathrm{t}$ and $A_\mathrm{b}$ as another metric
\begin{equation}
	m_\mathrm{lc} = \sqrt{\frac{1}{N_\mathrm{b}} \sum\limits_{(i,j)\in A_\mathrm{b}} (I_\mathrm{b}^{(i,j)}-\overline{I}_\mathrm{t})^2},
\end{equation}
where $\overline{I}_\mathrm{t}$ denotes the average intensity of the grayscale target area, and  $I_\mathrm{b}^{(i,j)}$ represents the pixel $I(i,j)$ in the local background area.
	
\textbf{Background complexity.} The scenes of MAVs can be diverse, such as skies, buildings, and mountains. The background complexity $m_\mathrm{bc}$ has a great influence on the detection results. $m_\mathrm{bc}$ is defined as
\begin{equation}
	m_\mathrm{bc} = \sqrt{\frac{1}{N_\mathrm{b}} \sum\limits_{(i,j)\in A_\mathrm{b}} (I_\mathrm{b}^{(i,j)}-\overline{I}_\mathrm{b})^2},
\end{equation}
where $\overline{I}_\mathrm{b}$ represents the average intensity of the local background area.

We use the difficulty matrix to classify the MAV targets into 4 types: Small Targets ($c_\mathrm{st} $), Low-Contrast images ($c_\mathrm{lc} $), Complex Backgrounds ($c_\mathrm{cb} $), and Simple Examples ($c_\mathrm{se}$). The classification criteria are as follows:

\begin{equation}
	\label{eq6}
	c_t=\left\{
	\begin{array}{lll}
		c_\mathrm{st}&, & m_\mathrm{ts} \leq \tau_{ts},  \\
		c_\mathrm{lc}&, & m_\mathrm{ts} > \tau_{ts} , m_\mathrm{lc} \leq \tau_{lc},\\
		c_\mathrm{se}&, & m_\mathrm{ts} > \tau_{ts}, m_\mathrm{lc} > \tau_{lc},m_\mathrm{bc} \leq\tau_\mathrm{bc}, \\
		c_\mathrm{cb}&, & m_\mathrm{ts} > \tau_{ts},m_\mathrm{lc} > \tau_{lc},m_\mathrm{bc} >\tau_\mathrm{bc},
	\end{array}
	\right.
\end{equation}
where $c_t$ represents the difficulty category of a target,  $\tau_\mathrm{ts}$, $\tau_\mathrm{lc}$ and  $\tau_\mathrm{bc}$ are the thresholds for difficulty matrix $m_\mathrm{ts}$, $m_\mathrm{lc}$ and  $m_\mathrm{bc}$, respectively. When the target has a small area, the target type is to be $c_\mathrm{st}$ without the consideration of the other two indexes. Thus there are four types of targets instead of eight types.
The thresholds $\tau_\mathrm{ts}$, $\tau_\mathrm{lc}$ and  $\tau_\mathrm{bc}$ are set to 16$\times$16, 10, and 10, respectively.
\begin{figure*}[t]
\centering
\includegraphics[width=1\textwidth]{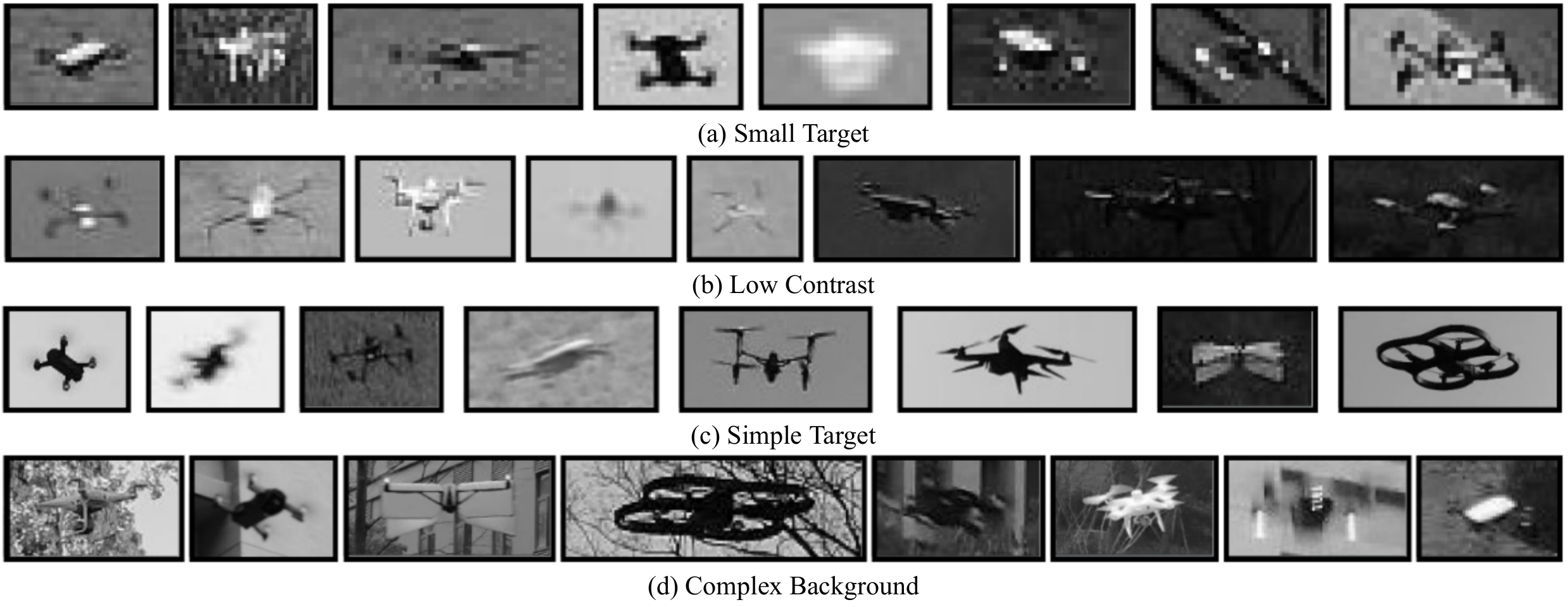} 
\caption{The four categories generated by the difficulty matrix from our M3D-Real dataset. Here, we use the grayscale image for the calculation of the four categories. Since the area of a small target is only 16$\times$16, the enlarged samples are not clear in the figure.}
\label{FourEvaluation}
\end{figure*}

Some examples of the classification results are shown in Fig.~\ref{FourEvaluation}. The MAVs in the category of small targets are more ambitious than other examples because they are extremely small objects in images. Compared to other categories, the examples in the low-contrast category are not easily discernible from the local backgrounds. The category of simple targets is the easiest to be detected. The samples have the simplest backgrounds and the highest contrast between the foregrounds and the backgrounds. The backgrounds of examples in the complex background category are more complex than other categories, leading to the detection difficulty.
\subsubsection{Adaptive threshold adjusting}
Pseudo-labels of different categories should be assigned adaptive thresholds instead of constant thresholds. Inspired by the work FlexMatch \cite{zhang2021flexmatch} which is used for semi-supervised image classification, we propose an adaptive threshold adjusting algorithm to tackle the cross-domain MAV detection issue. A set of candidate pseudo labels with confidence threshold $\tau_\mathrm{min}$ are separated into different difficulty categories. The relative difficulty $\sigma_t(c)$ for each category is defined as
\begin{equation}
	\sigma_t(c) = {\frac{1}{N_c}}{\sum\limits_{n=1}^{N_c}\mathbbm{1}( p_{y_n} > \tau_\mathrm{max})},
	\label{Threshold1}
\end{equation}
where $p_{y_n}$ is the output probability, $N_c$ represents the number of pseudo labels in category $c$ and $\tau_\mathrm{max}$ is the maximum confidence threshold. The adaptive threshold $\tau_t(c)$ is defined as
\begin{equation}
	\tau_t(c) = \max\left\{\frac{\sigma_t(c)}{\max \sigma_t}  \tau_\mathrm{max},\tau_\mathrm{min}\right\}.
	\label{Threshold2}
\end{equation}
After obtaining the threshold for each category, the pseudo labels will be corrected for training.

\subsection{Masked Copy-Paste Augmentation}
\label{sec:mca}
In this section, we propose a module called Masked Copy-paste Augmentation (MCA) to generate true labels on target images and reduce the noise caused by pseudo-labeling. Traditional copy-paste algorithms \cite{ghiasi2021simple, kisantal2019augmentation} need to use segmentation masks as one of the inputs and mainly focus on supervised object detection. There have not been copy-paste-based methods for unsupervised cross-domain object detection yet. Our masked copy-paste algorithm only requires bounding boxes of the targets instead of segmentation masks.

\begin{figure}[t]
	\centering
	\includegraphics[width=1\columnwidth]{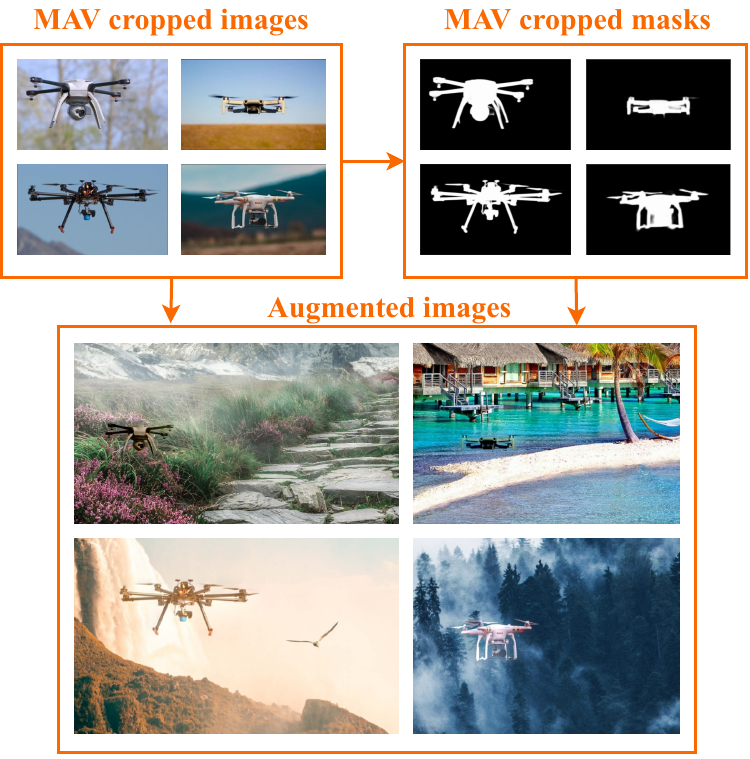} 
	\caption{Masked copy-paste augmentation algorithm.}
	\label{Copy-paste}
\end{figure}
Fig.~\ref{Copy-paste} illustrates the masked copy-paste augmentation algorithm. Since MAV targets are salient in the local areas, we use saliency segmentation \cite{qin2020u2} to obtain the segmentation masks of cropped MAV images. To guarantee the style consistency of MAV cropped images and target images, we use Poisson Merging \cite{perez2003poisson} to merge two images and generate harmonious augmented target images. Then the MAVs pasted on the target images have true labels. Instead of utilizing the cropped images in the source dataset, we collect a set of MAV images from the Internet, cause MAVs in these images are more salient to generate precise segmentation masks. The reason that we adopted the additional dataset instead of the source dataset is that the precise masks can help the network adapt to the target domain better.

The pipeline of the masked copy-paste augmentation module is shown in Algorithm \ref{CopyPasteAlgorithm}. The cropped MAV images with ground truth are randomly pasted on the unlabeled target images. Then the true labels can guide the detectors to identify MAVs from backgrounds. Besides, the true labels keep the same sizes as the pseudo labels. Thus the augmentation algorithm can also help the detectors adapt to the size distributions of MAVs in the target domain.
\begin{algorithm}[htbp]
	\caption{Masked Copy-Paste Augmentation Algorithm}
	\label{CopyPasteAlgorithm}
	\textbf{Input}: MAV cropped images $\mathcal{I}$ and bounding boxes, Target images $\mathcal{T}$ with pseudo labels $\mathcal{Y}^{p}_\mathcal{T}$ \\
	\textbf{Parameter}: Paste times $J$, the number $N_\mathcal{T}$ of target images with pseudo labels\\
	\textbf{Output}: Augmented target images $\mathcal{T}_A$ with true labels $\mathcal{Y}^{t}_\mathcal{T}$ and pseudo labels $\mathcal{Y}^{p}_\mathcal{T}$
	\begin{algorithmic}[1] 
		\STATE Segment MAV cropped masks of cropped images.
		\FOR {each $n \in [1,N_\mathcal{T}]$}
		\STATE Randomly choose one pseudo label in  $y_{\mathcal{T}_n}^p$; \\
		\STATE Randomly choose $J$ MAV cropped images;
		\FOR {each $j \in [1, J]$}
		\STATE Keep the size of $I_j$ consistent with $y_{\mathcal{T}_n}^p$; \\
		\STATE Randomly choose the center of $I_j$ in the image $\mathcal{T}_n$; \\
		\STATE Paste resized $I_j$ with its mask on $\mathcal{T}_n$;\\
		\STATE Calculate the new bounding box $y_j$ of $I_j$;\\
		\STATE Add the true label $y_j$ into $y_{\mathcal{T}_n}^t$ ; \\
		\ENDFOR
		\STATE Obtain augmented image $\mathcal{T}_{A_n}$ and its true labels $y_{\mathcal{T}_n}^t$; \\
		\ENDFOR
		\STATE \textbf{return} augmented target images $\mathcal{T}_A$ and true labels $y_{\mathcal{T}}^t$.
	\end{algorithmic}
\end{algorithm}

As shown in Fig.~\ref{Drone Adaptor}, the training loss $\mathcal{L}$ contains a supervised loss $ \mathcal{L}_s$ and an unsupervised loss $ \mathcal{L}_u$. After data augmentation, the labels of target data contain not only the pseudo labels but also the true labels. Then the training loss contains a new supervised loss $\mathcal{L}_t$ from truly-labeled MAVs on target images, which can be represented as
\begin{equation}
	\mathcal{L} = \mathcal{L}_s + \alpha \mathcal{L}_u + \beta \mathcal{L}_t,
\end{equation}
where $\alpha$ and $\beta$ are trade-off parameters. By doing this, the masked copy-paste augmentation module can significantly eliminate the pseudo-label noises and help the model learn the features between MAVs and backgrounds.

\section{Experiments}
\subsection{ Experimental Setup}
\label{tasks}
To verify the effectiveness of our method, the experiments are conducted on three types of domain adaptation tasks: simulation-to-real adaptation, cross-scene adaptation, and cross-camera adaptation, respectively. The comparison of these three tasks is shown in Table~\ref{Task}. Benefiting from the diversity of data, our M3D dataset is utilized to build three types of tasks. All the target domains only contain realistic images to meet the requirements for real-world applications.
\begin{table*}[htbp] \centering
	\renewcommand{\arraystretch}{1.15}
	\begin{tabular}{ccc}
		\hline \hline
		\textbf{Scenarios} & \textbf{Datasets} & \textbf{Image Numbers} \\\hline
		Simulation vs Real  & M3D-Sim-Normal vs M3D-Real-Target & 18,476 vs 15,326 images \\
		Different scenes & M3D-Real-Source vs M3D-Real-Target & 39,933 vs 15,326 images\\
		Different cameras & M3D-Real vs Drone-vs-Bird & 55,259 vs 19,027 images \\
		\hline \hline
	\end{tabular}
	\caption{The comparison of different tasks in our benchmark.}
	\label{Task}
\end{table*}
\subsubsection{Simulation-to-real adaptation benchmark}
The M3D-Sim-Normal subset and M3D-Real-Target subset are chosen to study the simulation-to-real adaptation problem. The huge style and scene discrepancies between the simulation and the realistic domain can effectively validate the adaptation methods. Both the source dataset and the target dataset contain a large amount of data and challenging scenes.
\subsubsection{Cross-scene adaptation benchmark}
In this benchmark, we choose the M3D-Real-Source subset and the M3D-Real-Target subset as the source and the target domain. The source images and the target images are captured by the same camera. The only difference between these two sets is the backgrounds of MAVs. Thus this task can help to verify the cross-scene adaptation without the perturbation of the devices.
\subsubsection{Cross-camera adaptation benchmark}
Drone-vs-Bird dataset \cite{coluccia2021drone} mainly focuses on small MAV detection but with a fixed camera. Due to the difference between the M3D dataset and the Drone-vs-Bird dataset, we choose these two datasets to study the cross-camera adaptation (M3D-Real$\rightarrow$Drone-vs-Bird) problem. The videos in this dataset are extracted to a set of images every 5 frames and randomly separated into the training set, validation set, and testing set, which contain 13,318, 1,902, and 3,807 images, respectively.

These three tasks are challenging and worthy of study.

\subsection{Implementation Details}
Following the work \cite{ramamonjison2021simrod, mattolin2023confmix}, we choose YOLOv5 as the backbone model. We choose the YOLOv5x as the large model and the YOLOv5s as the small model. The images are resized to 640$\times$640 before training. The learning rate for the source model is set at 0.01. During the whole adaptive training procedure, the learning rate is set to 0.002.
Both the large and small source models are trained for 50 epochs with a batch size of 32 on 4 GPUs and 1 GPU, respectively. During the large-to-small model training process, the detection model is trained for 30 epochs at each training period. Each target image is pasted with truly-labeled MAVs 3 times ($J=3$). The setting is similar to the result reported in \cite{rui2021comprehensive}. The width $w_{\mathrm{b}}$ and the height $h_{\mathrm{b}}$ of the local background area are chosen to be 1.5 times the width $w_{\mathrm{t}}$ and the height $h_{\mathrm{t}}$ of the target area. 

We use mAP(\%) as the main evaluation metric when $\mathrm{IOU} = 0.5$. We also compare the detection results with the Source Only way (trained on source domain) and the Oracle way (trained on target domain) to evaluate the improved performance of domain adaptation algorithms. Thus we adopt the Adaptation Gain $\rho$ in \cite{ramamonjison2021simrod} as the quantitative evaluation index. The calculation of $\rho$ is represented as
\begin{equation}
 \rho = \frac{\mathrm{mAP}(\theta^a) - \mathrm{mAP}(\theta^s)}{\mathrm{mAP}(\theta^o) - \mathrm{mAP}(\theta^s)} \times 100\%,
\end{equation}
where $\theta^a$, $\theta^s$, and $\theta^o$ represent the adaptation model, Source Only model, and the Oracle model, respectively. The Source Only method means that the detection model is only trained on the source dataset. Oracle method represents the model trained on the target dataset.
\subsection{Main Results}
The main results of the three adaptation tasks are shown as follows.  NSN \emph{ w/o teacher} represents the detection results of our NSN method without the large model.
\subsubsection{Simulation-to-real adaptation}

We report the simulation-to-real adaptation results in Table~\ref{Exp1}. Benefiting from the diversity of our simulation dataset, we can observe that the Source Only model achieves 39.9\% mAP, which is much higher than the Source Only model trained by the M3D-Real-Source subset. However, the baseline method SimROD only improves the detection performance by 1.2\% due to the huge domain gap between the simulation and real datasets.  In particular, our NSN model exceeds SimROD by 5.8\% on the task of simulation-to-real adaptation. This demonstrates the effectiveness of our prior-guided curriculum learning and masked copy-paste augmentation modules. 

From Table~\ref{Exp1}, we can find that most state-of-the-art methods perform even poorer than the Source Only method. This phenomenon does not happen in the other two adaptation tasks. The reason for degradation is as follows.

First, the self-training-based domain adaptation methods are easily affected by the pseudo-label noises. The inaccuracy of the pseudo labels will increase when meeting smaller MAVs under more diverse backgrounds. Due to the large gap between the simulation domain and the reality domain, the noises of the pseudo labels are more challenging in this scenario. If the noises exceed a certain level, the model will get worse and worse as the training continues. Therefore, the influence of noises makes the detection networks confused and causes poor detection performances. As shown in Table~\ref{Exp1}, many methods obtain poorer performances than the Source Only method.
Second, the SimROD method tries to solve this problem by introducing a teacher-student model to take advantage of the large model to reduce the noise. Thus, the SimROD method performs better than the SimROD w/o teacher method. However, the improvement ability is still limited because the pseudo-label noises are more complex when the targets are MAVs.
Finally, the reason that we designed the noise suppression network is to solve this problem. We tackle this problem from two approaches. On the one hand, we use the masked copy-paste augmentation module to help the detection model fully adapt to the backgrounds in the target domain. On the other hand, we use the prior-guided curriculum learning module to improve the precision of the pseudo labels. 

\begin{table}[htbp] \centering
	\renewcommand{\arraystretch}{1.3}
	\begin{tabular}{l|c|c|r}
		\hline \hline
		\textbf{Methods} & \textbf{Backbone} & \textbf{mAP (\%)} & \multicolumn{1}{c}{\textbf{$\rho$}} \\\hline
		Source Only & Yolov5s & 39.9 & 0\% \\\hline
            CSL \cite{soviany2021curriculum} & Yolov5s & 24.6 & -30.9\% \\
            AsyFOD \cite{gao2023asyfod} & Yolov5x & 35.4 & -9.1\%\\
            AcroFOD \cite{gao2022acrofod} & Yolov5x &21.9 & -36.3\%\\
		ConfMix \cite{mattolin2023confmix} & Yolov5s & 37.4 & -5.0\%  \\
		SimROD \cite{ramamonjison2021simrod} & Yolov5s & 41.1 & 2.4\%  \\
		SimROD \emph{w/o teacher} & Yolov5s & 38.5 & -2.8\% \\\hline
		NSN \emph{w/o teacher} & Yolov5s & 41.2 & 2.6\%  \\
		NSN (Ours)   & Yolov5s  & \textbf{46.9} & \textbf{14.1\%} \\\hline
		Oracle & Yolov5s &  89.5 & 100\%\\
		\hline \hline
	\end{tabular}
	\caption{The \textbf{simulation-to-real adaptation} performance of different methods.}
	\label{Exp1}
\end{table}

\subsubsection{Cross-scene adaptation}
The cross-scene adaptation results are shown in Table~\ref{Exp2}. The M3D-Real dataset contains plenty of scenes and is captured by a moving camera. Even if there is no cross-device problem to be solved, the cross-scene adaptation of the M3D dataset is still a challenging issue. The Source Only model only achieves 28.7\% mAP on the target dataset, which indicates that the MAV detection model suffers from significant domain discrepancy. The detection improvement of our method is 21.8\%. Among all the comparison methods, AsyFOD \cite{gao2023asyfod} has the best performance and achieves 47.9\% mAP on this task, which is still lower than our proposed method. Our NSN method exceeds the baseline algorithm of 3.7\% on the task of cross-scene adaptation. Furthermore, the simulation-to-real adaptation and cross-scene adaptation tasks contain the same target dataset M3D-Real-Target, thus we compare the detection results of these two tasks. We can observe that even though the Source Only model of cross-scene adaptation is lower, the final adaptation result and adaptation gain are much higher than the simulation-to-real adaptation. The reason is that the pseudo-label noises are more influenced by the huge style differences in the simulation-to-real task and make the detection model more confused. Our method can perform better on the simulation-to-real task when combined with adversarial training modules. In this paper, we pay more attention to the real-to-real issue.
\begin{table}[htbp] \centering
	\renewcommand{\arraystretch}{1.3}
	\begin{tabular}{l|c|c|r}
		\hline \hline
		\textbf{Methods} & \textbf{Backbone} & \textbf{mAP (\%)} & \multicolumn{1}{c}{\textbf{$\rho$}} \\\hline
		Source Only & Yolov5s & 28.7 & 0\% \\\hline
            CSL \cite{soviany2021curriculum} & Yolov5s & 34.4 & 9.3\%\\
            AsyFOD \cite{gao2023asyfod} & Yolov5x &  47.9 & 31.6\% \\
            AcroFOD \cite{gao2022acrofod} & Yolov5x & 30.4 & 2.8\%\\
		ConfMix \cite{mattolin2023confmix} & Yolov5s & 32.2 & 5.8\%\\
		SimROD \cite{ramamonjison2021simrod} & Yolov5s &  46.8 & 29.8\% \\
		SimROD \emph{w/o teacher} & Yolov5s & 39.1 & 17.1\% \\\hline
		NSN \emph{w/o teacher} & Yolov5s & 43.1 &  23.7\%\\
		NSN (Ours)   & Yolov5s  & \textbf{50.5} & \textbf{35.9\%}  \\\hline
		Oracle & Yolov5s &  89.5 & 100\% \\
		\hline \hline
	\end{tabular}
	\caption{The \textbf{cross-scene adaptation} performance of different methods.}
	\label{Exp2}
\end{table}
\subsubsection{Cross-camera adaptation}
There are huge differences in the target size and backgrounds of images between the M3D-Real dataset and the Drone-vs-Bird dataset. The cross-camera adaptation performance of different methods is shown in Table~\ref{Exp3}. Thus the detection result is low when only using the source data, which is only 31.9\%. Compared to the SimROD algorithm, our method achieves a huge improvement of 11.3\% on the task of cross-camera adaptation. The performance of the small model is significantly improved by 18.9\% with the help of the large model. Our method achieves higher adaptation gain than other tasks because some images in the Drone-vs-Bird dataset are shot under a still camera, thus the detection model can adapt to the new environments faster and perform better on the target domain. 

\begin{table}[htbp] \centering
	\renewcommand{\arraystretch}{1.3}
	\begin{tabular}{l|c|c|r}
		\hline \hline
		\textbf{Methods} & \textbf{Backbone} & \textbf{mAP (\%)} & \multicolumn{1}{c}{\textbf{$\rho$}} \\\hline
		Source Only & Yolov5s & 31.9 & 0\% \\\hline
            CSL \cite{soviany2021curriculum} & Yolov5s& 33.9 &3.5\%\\
            AsyFOD \cite{gao2023asyfod} & Yolov5x& 39.8& 13.8\%\\
            AcroFOD \cite{gao2022acrofod} & Yolov5x & 35.7 & 6.6\%\\
		ConfMix \cite{mattolin2023confmix} & Yolov5s & 25.8& -10.7\% \\
		SimROD \cite{ramamonjison2021simrod} & Yolov5s & 50.2 & 32.0\%\\
		SimROD \emph{w/o teacher} & Yolov5s & 36.5 & 8.0\% \\\hline
		NSN \emph{w/o teacher} & Yolov5s & 42.6 & 18.7\% \\
		NSN (Ours) & Yolov5s & \textbf{61.5} & \textbf{51.8\%} \\\hline
		Oracle & Yolov5s & 89.1 & 100\% \\
		\hline \hline
	\end{tabular}
	\caption{The \textbf{cross-camera adaptation} performance of different methods.}
	\label{Exp3}
\end{table}

After comparing the results of these three tasks, we can conclude that the pseudo-labeling technique is more suitable for dealing with scene-to-scene and real-to-real adaptation issues. Our method can significantly reduce the pseudo-label noises and increase the adaptation performance.
\subsection{Qualitative Analysis}
\begin{figure*}[t]
	\centering
	\subfigure[Source Only]{
		\centering
		\includegraphics[height=2in,width=2.2in]{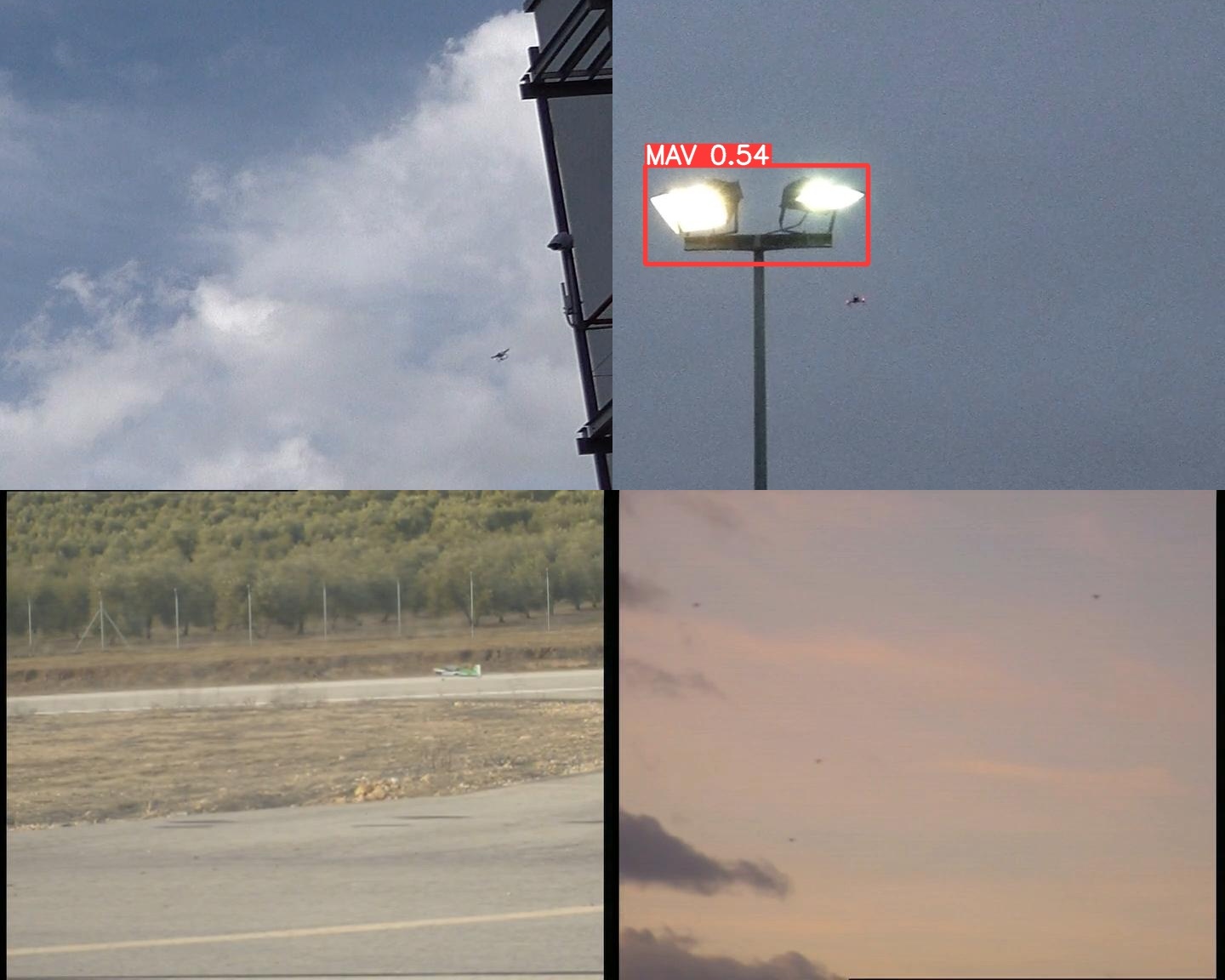}
	}
	\centering
	\subfigure[SimROD]{
		\centering
		\includegraphics[height=2in,width=2.2in]{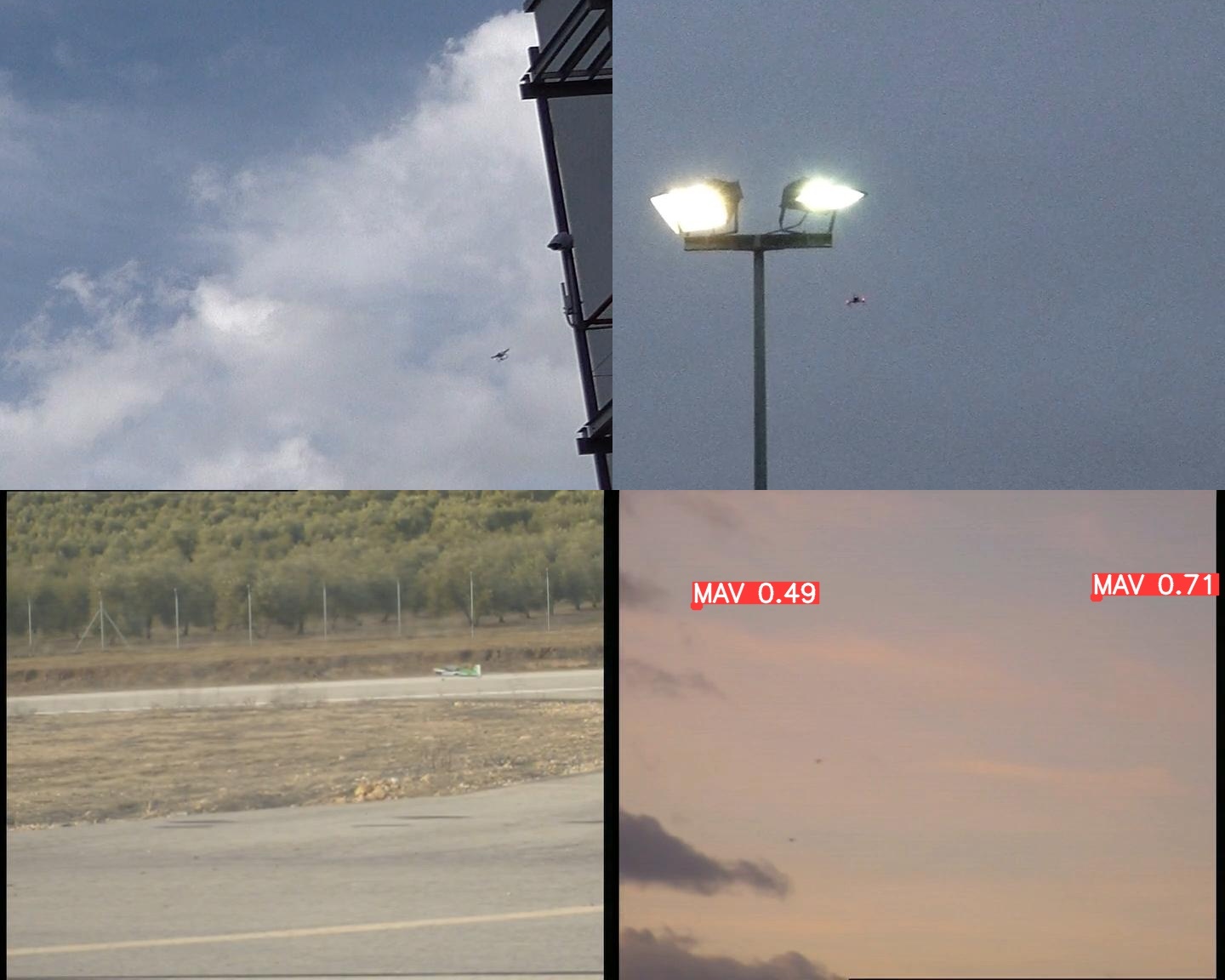}
	}
	\centering
	\subfigure[NSN (Ours)]{
		\centering
		\includegraphics[height=2in,width=2.2in]{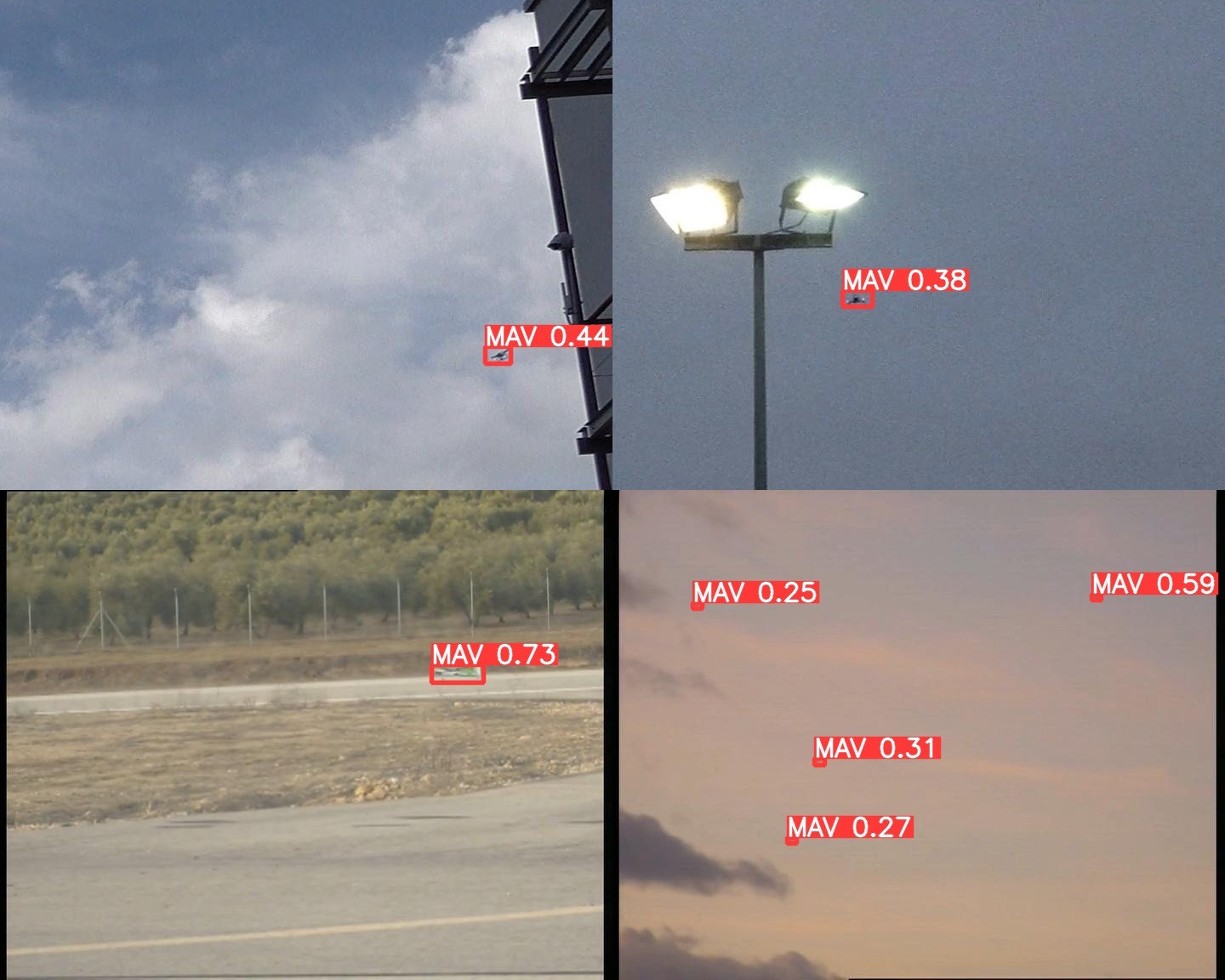}
	}
	\centering
	\caption{The qualitative comparison of different methods on the cross-camera adaptation task.}
	\label{Visualization}
\end{figure*}
The qualitative comparison of different methods is shown in Fig~\ref{Visualization}. The detection results on target testing examples in the Drone-vs-Bird dataset are presented in the figure. The comparison consists of three methods, which are Source Only, SimROD, and NSN (Ours) methods. When only the Source Only method is adopted, the detection model cannot detect MAVs and wrongly classify other objects as MAVs. Even though the baseline method SimROD detects some MAVs in the images, there are still some tiny MAVs not detected. Our method significantly improves the detection performance of missed targets with the help of true labels generated by our masked copy-paste augmentation algorithm. Furthermore, our method can also help to reduce the wrong detection rate benefit of the PCL module.
\subsection{Ablation Study}
\subsubsection{Ablation study of different modules}
\begin{table}[htbp] \centering
	\renewcommand{\arraystretch}{1.3}
	\begin{tabular}{c|ccc|c|r}
		\hline \hline
		\textbf{Method} & \textbf{LM} & \textbf{PCL} &\textbf{MCA} & \textbf{mAP (\%)} & \multicolumn{1}{c}{\textbf{$\rho$}} \\
		\hline
		Source Only  & -& -& -& 31.9& 0\%\\\hline
		\multirow{5}{*}{Proposed} & & \checkmark &  &  37.5 & 9.8\% \\
		 & &  & \checkmark &  41.1 & 16.1\% \\
		& & \checkmark & \checkmark &  42.6 & 18.7\%\\
		& \checkmark &  & &  50.2 & 32.0\% \\
		& \checkmark &  & \checkmark &  60.5 & 50.0\%\\\hline
		Ours & \checkmark & \checkmark & \checkmark & \textbf{61.5} & \textbf{51.8\%} \\
		\hline \hline
	\end{tabular}
	%\end{tabularx}
	\caption{Ablation study on the task of cross-camera adaptation. The LM, PCL, and MCA represent the Large Model, Prior-guided Curriculum Learning module, and Masked Copy-paste Augmentation module, respectively. }
	\label{AblationStudy}
\end{table}
The ablation study is mainly conducted on the task of cross-camera adaptation. Table \ref{AblationStudy} shows a comprehensive comparison of the contributions of different modules in our NSN method. First, with the participation of the large model, the improvement of the adaptive small detection model is further increased from 10.7\% to 29.6\%. Second, the MCA module improves the performance of the small model by 9.2\%. After the small model owns the knowledge of the large model, the MCA module still contributes the improvement of 10.3\%. Finally, the PCL module is also indispensable. For example, if we skip this module, the performance of the purely small model drops by 1.5\%, and our NSN method drops by 1.0\%. The integration of these modules is more efficient than individual modules. 

\subsubsection{The choice of cropped images for data augmentation}
In the masked copy-paste augmentation module, we use an additional saliency dataset with 2,113 images as cropped images, which can be downloaded from our GitHub. Some examples are shown in Fig.~\ref{Copy-paste}. Compared with the examples in the source dataset (see Fig.~\ref{FourEvaluation}), the MAVs in the saliency dataset are bigger and more salient enough for saliency segmentation. We conduct two experiments to show the design of this module. 

First, we compare our method using the additional salient images with the cropped images whose widths exceed 50 pixels from the source dataset. To reduce the influence of other modules, we only adopt the MCA module for comparison. The detection results are shown in Table~\ref{Exp-copypaste}. When we use the cropped images directly from the source dataset, the adaptation achieves 40.6\% mAP, which is smaller than the training with the saliency dataset. The reason is that the segmentation masks of small objects are not easily obtained automatically from the saliency segmentation algorithm. If we use all the images from the source dataset, the detection results will drop deeply due to the inaccurate masks. In conclusion, when the MAVs in the source dataset are not big and salient enough for saliency segmentation, it is suggested to choose the additional dataset for utilization.  Our proposed masked copy-paste module could work well whether the cropped images come from the source dataset or an additional dataset as long as we can obtain the precise masks of the targets. 

Second, to guarantee fairness, we also show the experimental results of SimROD when expanding the source dataset with the additional salient dataset during training. As shown in Table~\ref{Exp-copypaste}, the detection performance drops to 35.5\% when more salient images are added into the training period. The reason is that the MAVs in the salient images take up a large area. The distribution of the saliency dataset is different from the source dataset and the target dataset. The additional salient dataset can only play a significant role in our masked copy-paste module.
\begin{table}[htbp] \centering
	\renewcommand{\arraystretch}{1.3}
	\begin{tabular}{l|c|c|r}
		\hline \hline
		\textbf{Methods} & \textbf{Backbone} & \textbf{mAP (\%)} & \multicolumn{1}{c}{\textbf{$\rho$}} \\\hline
		Source Only & Yolov5s & 31.9 & 0\% \\\hline
  		MCA (Saliency) & Yolov5s & \textbf{41.1} & \textbf{16.1\%} \\
            MCA (Source, $w_t$ $\geq$ 50 pixels) & Yolov5s & 40.6 & 15.2\% \\\hline
		SimROD \emph{w/o teacher} & Yolov5s & 36.5 & 8.0\% \\
		SimROD \emph{w/o teacher} + Saliency & Yolov5s & 35.5 & 6.3\% \\\hline
		Oracle & Yolov5s & 89.1 & 100\% \\
		\hline \hline
	\end{tabular}
	\caption{The ablation study of the masked copy-paste module.}
	\label{Exp-copypaste}
\end{table}

\subsubsection{Varying of different hyper-parameters} The proposed modules contain a few hyper-parameters that could influence the detection results. We conduct two experiments to show the effect of different values of the $J$ and $\tau_{max}$. To ensure that the results are not disturbed by the participation of other modules, the experiments are independently conducted on the modules themselves. The detection results are shown in Table~\ref{Hyper}.

When we change the times of the copy-paste module, the detection model achieves the highest result when $J$ equals three. This phenomenon is consistent with the experiment results reported in \cite{kisantal2019augmentation}. The reason is that when the copy-paste time is too high, the detection is easily overfitting. We also conduct experiments of varying the value of $\tau_{max}$. The detection results in Table~\ref{Hyper} show that the PCL module can achieve the best performance when $\tau_{max}$ equals $0.75$. As for another hyper-parameter $\tau_{min}$, we follow the default setting $0.25$ of the Yolov5 network for stable performances in different scenarios.

\begin{table}[htbp] \centering
	\renewcommand{\arraystretch}{1.3}
	\begin{tabular}{l|c|c|r}
		\hline \hline
		\textbf{Method} & \textbf{Hyper-parameters} & \textbf{mAP (\%)} & \multicolumn{1}{c}{\textbf{$\rho$}} \\\hline
		Source Only & - & 31.9 & 0\% \\\hline
		  MCA & $J=1$ & 37.4 & 9.6\% \\
            MCA & $J=3$ & \textbf{41.1} & \textbf{16.1\%} \\
		MCA & $J=5$ & 39.8 & 13.8\% \\\hline
		PCL& $\tau_{max}=0.6$ & 37.2 & 9.3\% \\
            PCL& $\tau_{max}=0.75$ & \textbf{37.5} & \textbf{9.8\%} \\
		PCL& $\tau_{max}=0.9$ & 36.7 & 8.4\% \\ \hline
		Oracle & - & 89.1 & 100\% \\
		\hline \hline
	\end{tabular}
	\caption{The experiments of varying the values of different hyper-parameters on the task of cross-camera adaptation.}
	\label{Hyper}
\end{table}

\subsection{Experimental Analysis}
\subsubsection{The influence of adaptive thresholds}
\begin{figure}[t]
\centering
\includegraphics[width=1\columnwidth]{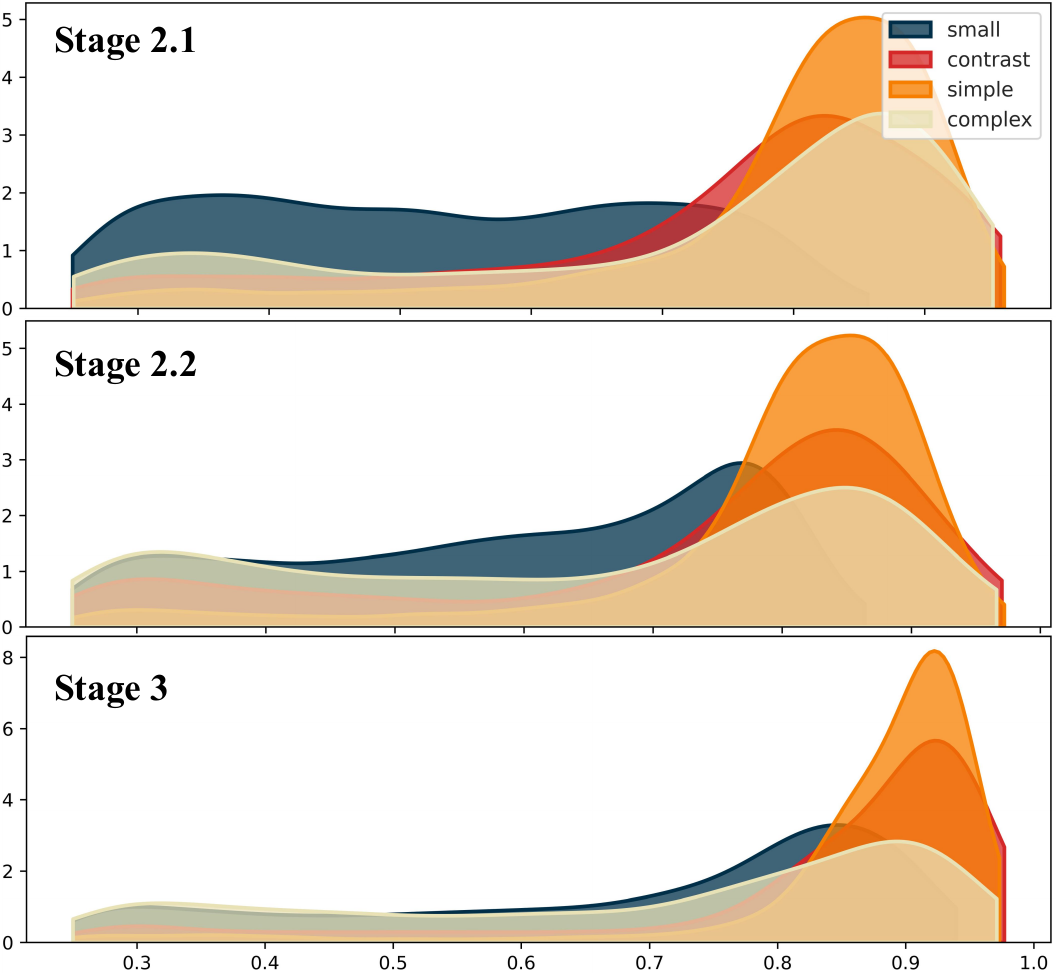} 
\caption{The confidence distribution of pseudo labels from different categories during the training process.}
\label{Distribution}
\end{figure}

In this section, we study the influence of adaptive thresholds on pseudo labels. Furthermore, we use Fig.~\ref{Distribution} to prove the effectiveness of the PCL module and how it influences the choice of pseudo labels at different training stages. The confidences of pseudo labels from different difficulty categories show different distributions. The simple category tends to have pseudo labels of high confidence. As the progress of training, the confidences of pseudo labels become larger and all the distributions move towards the right side of the figure. The adaptive thresholds generated by our PCL module also gradually increase to choose more reliable pseudo labels.

\subsubsection{The effectiveness of the data augmentation}
\begin{figure}[t]
	\centering
	\subfigure[CutMix]{
		\centering
		\includegraphics[width=0.46\columnwidth]{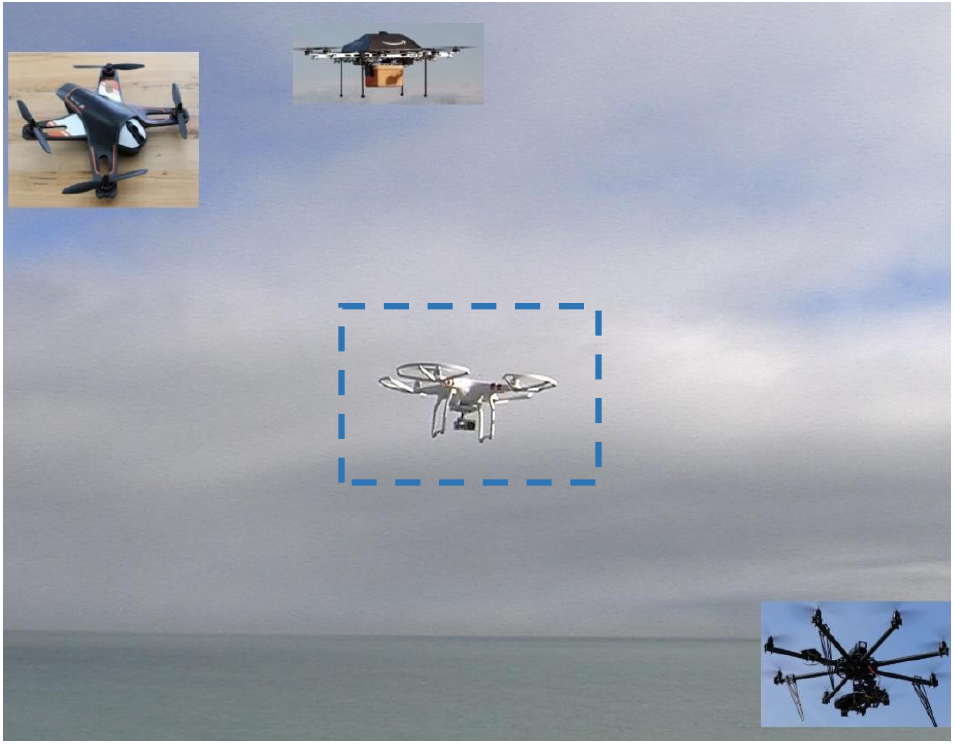}
	}
	\centering
	\subfigure[MCA (Ours)]{
		\centering
		\includegraphics[width=0.46\columnwidth]{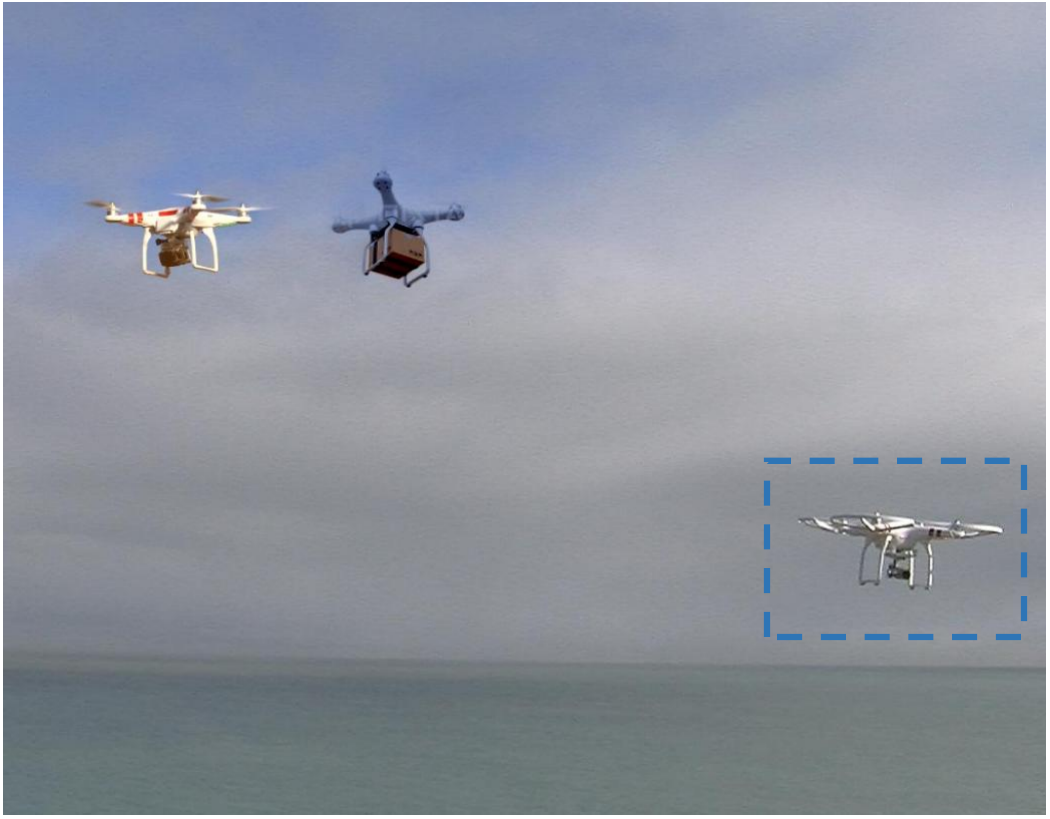}
	}
	\centering
	\caption{The augmentation results of CutMix and MCA (Ours) on Drone-vs-Bird dataset. Original images only contain one MAV which is enclosed within a blue box. }
	\label{Aug}
\end{figure}

To show the effectiveness of our data augmentation module, we also compare our method with other data augmentation methods. To reduce the influence of other modules in our method, we only adopt the MCA module in the experiment. The results are shown in Table~\ref{AugComparison}. When we replace our MCA module with CutMix, the detection result drops from 41.1\% to 33.0\%, proving the significance of our data augmentation module. The reason is that the local background doesn't change when pasting the cropped MAV images on the target images directly. The differences between CutMix and MCA are shown in Fig.~\ref{Aug}. Our augmentation module can generate more harmonious results. The detection model could not learn the local background features of target images when using CutMix augmentation.
\begin{table}[htbp] \centering
	\renewcommand{\arraystretch}{1.3}
	\begin{tabular}{l|c|c|r}
		\hline \hline
		\textbf{Methods} & \textbf{Backbone} & \textbf{mAP (\%)} & \multicolumn{1}{c}{\textbf{$\rho$}} \\\hline
		Source Only & Yolov5s & 31.9 & 0\% \\\hline
		MCA (Ours) & Yolov5s & \textbf{41.1} & \textbf{16.1\%} \\
		ConfMix \cite{mattolin2023confmix} & Yolov5s & 25.8 &-10.7\%\\
		CutMix \cite{yun2019cutmix} & Yolov5s & 33.0 & 1.9\% \\\hline
		Oracle & Yolov5s & 89.1 & 100\% \\
		\hline \hline
	\end{tabular}
	\caption{The cross-camera adaptation performance of different augmentation methods.}
	\label{AugComparison}
\end{table}

\subsubsection{Texture randomization}
\label{sec:texture}
\begin{figure}[t]
	\centering
	\includegraphics[width=1\columnwidth]{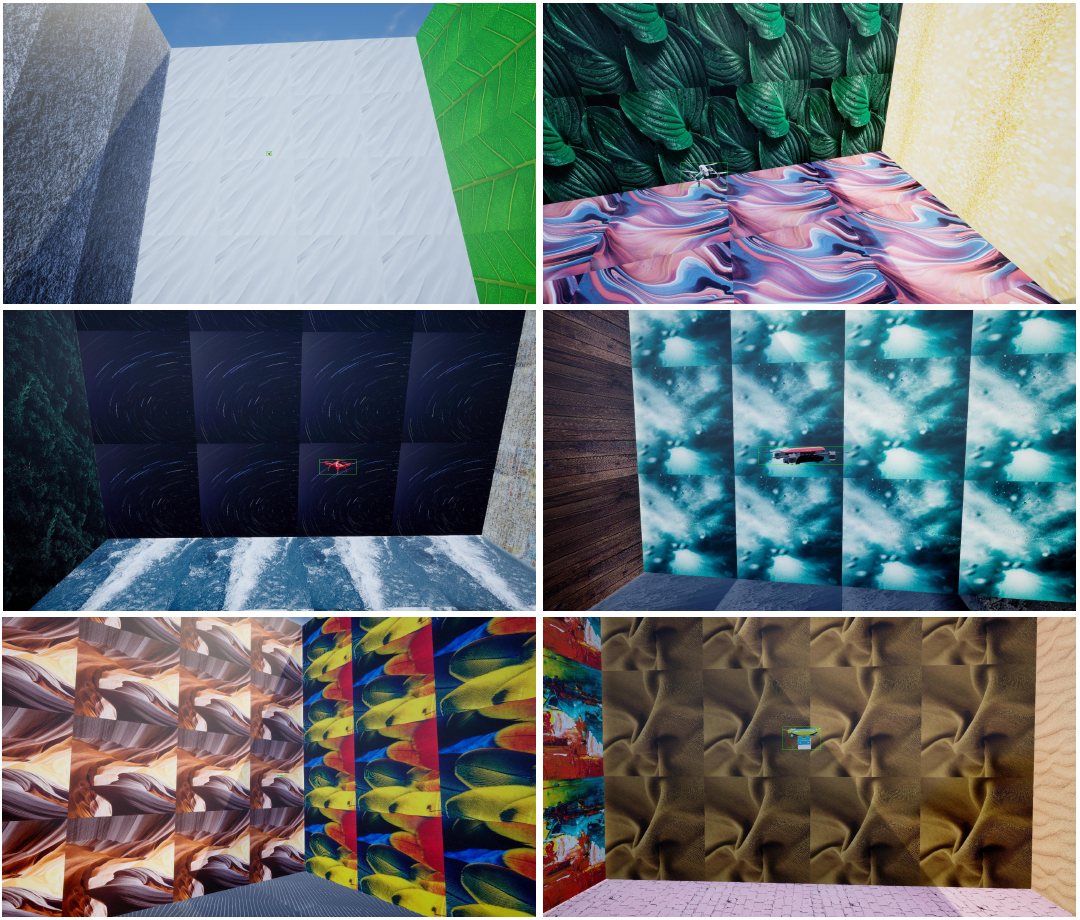} 
	\caption{The images generated by texture randomization in M3D-Sim subset. The walls in Unreal Engine dynamically change their textures during data collection.}
	\label{Texture}
\end{figure}

\begin{table}[htbp] \centering
	\renewcommand{\arraystretch}{1.3}
	\begin{tabular}{l|c|c}
		\hline \hline
		\textbf{Dataset} & \textbf{Backbone} & \textbf{mAP (\%)} \\\hline
		M3D-Sim-Normal & Yolov5s & 39.9 \\
		M3D-Sim-Texture & Yolov5s & 22.2 \\
		M3D-Sim & Yolov5s & \textbf{43.5} \\ 
		\hline \hline
	\end{tabular}
	\caption{The detection performances trained on different subsets in the M3D-Sim dataset.}
	\label{TextureRandomExp}
\end{table}
M3D-Sim subset contains normal images and images generated by texture randomization. The visual differences between these two types of images can be seen in Fig.~\ref{Airsim} and Fig.~\ref{Texture}. Texture randomization is one way to reduce the gap between simulation and reality, thus we also conduct experiments to verify the effectiveness of this approach. The results are shown in Table~\ref{TextureRandomExp}. We train the detection model by only using the normal data, randomized data, and all the data, respectively. The model is tested on the M3D-Real-Target dataset. The results of these three approaches are 39.9\%, 22.2\%, and 43.5\%, respectively.  It can be concluded that texture randomization can indeed improve detection performance in a supervised way, but its ability is limited. Besides, this type of method can improve the generalization ability of the detection model but cannot help the model adapt to new environments.

\subsubsection{Computational cost} The computational cost can be divided into the training period and the inference period. First, the training time of our method takes a few hours more than the baseline SimROD method. Take the task of cross-camera adaptation as an example. The training of the SimROD method takes 45 hours and our method costs 47 hours. The additional two hours are cost by the MCA and the PCL module. Second, the baseline SimROD method and our NSN method contain the same detection model, thus the inference time is the same as the Yolov5s model.

\subsection{The Universal MAV Detection Model based on M3D dataset}
As a dataset containing diverse images, the M3D dataset can be chosen as the common dataset for MAV detection. This section shows the experiment results when using different training subsets of M3D for training. All the experiments still choose the Yolov5s network as the MAV detection model. The models are tested on the testing set of M3D-Real-Source, M3D-Real-Target, and M3D-Real, respectively. When we use the whole training set of the M3D dataset as the training dataset, the detection model achieves 88.3\% mAP on the testing set of M3D-Real. The results can serve as the precision of the Oracle model for study in the future. MAV detection is critical for real-world applications. Thus the results can also be utilized as the universal MAV detection model for the community.
\begin{table}[htbp] \centering
	\renewcommand{\arraystretch}{1.3}
	\begin{tabular}{c|c|c|c}
		\hline \hline
		\textbf{Training Subset} & \textbf{M3D-Real-S} & \textbf{M3D-Real-T} & \textbf{M3D-Real}\\\hline
            M3D-Sim-Normal & 53.5\% & 39.9\% & 50.6\%\\
		M3D-Sim-Texture & 38.4\% & 22.2\%  & 34.6\%\\ \hline
            M3D-Real-Source & 86.2\% & 28.7\% & 73.4\%\\
		M3D-Real-Target & 46.1\% & 89.1\%  & 56.1\%\\ \hline
		M3D-Sim & 57.0\% & 43.5\% & 54.2\%\\
		M3D-Real & 88.2\% & 90.0\%  & 88.6\%\\ \hline
		M3D & 88.2\% & 89.0\% & 88.3\%\\ 
		\hline \hline
	\end{tabular}
	\caption{The detection performances trained on different training subsets of the M3D dataset. The M3D-Real-S and M3D-Real-T represent the M3D-Real-Source and M3D-Real-Target subsets, respectively. The detection results are reported on the testing subsets.}
	\label{WholeMAVTest}
\end{table}

\section{Conclusion}
This paper benchmarks the cross-domain MAV detection problem. We first propose a Multi-MAV-Multi-Domain (M3D) dataset and construct a novel domain adaptive MAV detection benchmark consisting of three representative domain adaptation tasks, i.e., simulation-to-real adaptation, cross-scene adaptation, and cross-camera adaptation. Moreover, we propose a novel noise suppression network with a prior-guided curriculum learning module, a masked copy-paste augmentation module, and a large-to-small model training procedure. Our method significantly reduces the pseudo noises and achieves real-time MAV detection. The results of extensive experiments and ablation studies demonstrate the effectiveness of the proposed method. We have released the dataset, hoping our study could advance the cross-domain MAV detection research.

Due to the large domain shift of MAV detection, there is still a huge improving space to achieve the performance of supervised learning. We will consider the spatial and temporal information in future work.

\bibliography{zsyReferenceAll} 
\bibliographystyle{ieeetr}

%========================================================================
\begin{IEEEbiography}[{\includegraphics[width=1in,height=1.25in,clip,keepaspectratio]{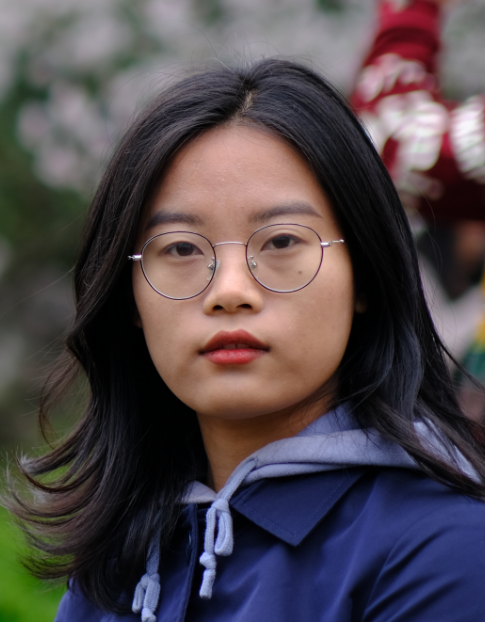}}]
{Yin Zhang} received the B.Sc. degree in measurement and control technology and instrumentation from Tianjin University, Tianjin, China, in 2017, and an M.Sc. degree in instrument science and technology from BUAA University, Beijing, China, in 2020. She is currently working toward a Ph.D. degree with the Intelligent Unmanned Systems Laboratory, at Westlake University, Hangzhou, China.
Her research interests include domain adaptation, MAV detection, and depth estimation.	
\end{IEEEbiography}
\begin{IEEEbiography}[{\includegraphics[width=1in,height=1.25in,clip,keepaspectratio]{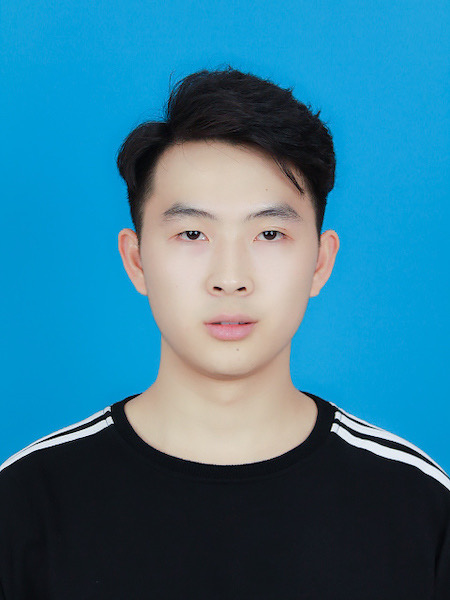}}]
{Jinhong Deng} received the B.Eng. degree from the Southwest Jiaotong University, Chengdu, China in 2019. He is currently pursuing the Ph.D. degree with the School of Computer Science and Engineering, University of Electronic Science and Technology of China (UESTC). His research interests include transfer learning, label-efficient learning, and their applications in computer vision.	
\end{IEEEbiography}
\begin{IEEEbiography}[{\includegraphics[width=1in,height=1.25in,clip,keepaspectratio]{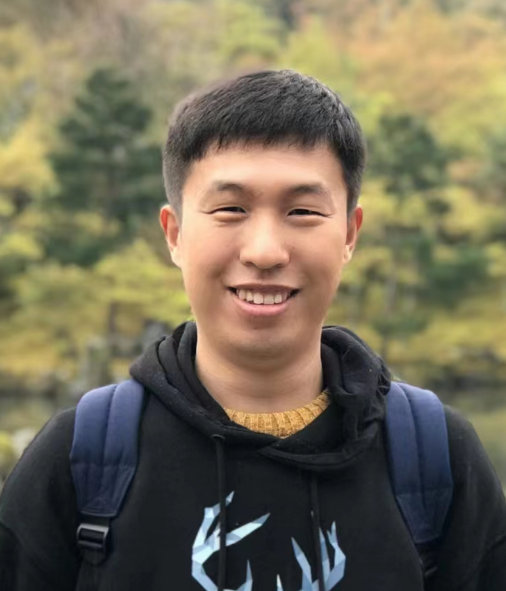}}]
{Peidong Liu}  received the B.E. and M.E. degrees from the National University of Singapore, in 2012 and 2015, respectively. He then joined ETH Zurich as a postgraduate student and received his PhD degree in Computer Science in 2021. He is currently an Assistant Professor with the School of Engineering, Westlake University, Hangzhou, China. His research interests lie in 3D computer vision and robotics.
\end{IEEEbiography}
\begin{IEEEbiography}[{\includegraphics[width=1in,height=1.25in,clip,keepaspectratio]{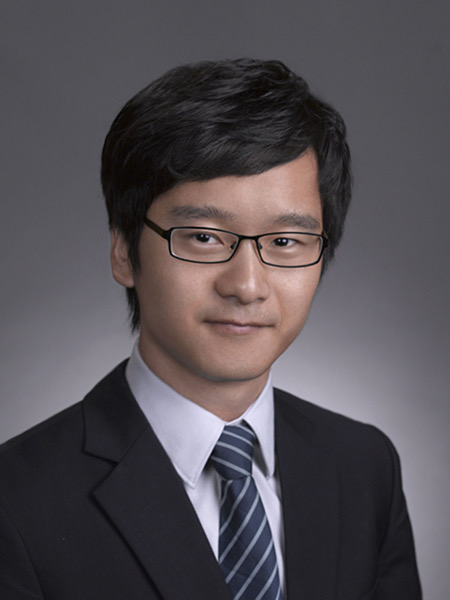}}]
{Wen Li} received the B.E. and M.S. degrees from the Beijing Normal University in 2007 and 2010, respectively, and the Ph.D. degree from the Nanyang Technological University in 2015.
He currently is a Full Professor at the School of Computer Science and Engineering, University of Electronic Science and Technology of China. His main research interests include transfer learning, multi-view learning, multiple kernel learning, and their applications in computer vision.
\end{IEEEbiography}

\begin{IEEEbiography}[{\includegraphics[width=1in,height=1.25in,clip,keepaspectratio]{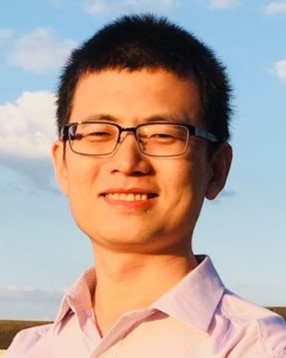}}]
{Shiyu~Zhao} received the B.Eng. and M.Eng. degrees from Beijing University of Aeronautics and Astronautics, Beijing, China, in 2006 and 2009, respectively, and the Ph.D. degree from the National University of Singapore, Singapore, in 2014, all in electrical engineering.

From 2014 to 2016, he was a Postdoctoral Researcher at the Technion-Israel Institute of Technology, Haifa, Israel, and the University of California at Riverside, Riverside, CA, USA. From 2016 to 2018, he was a Lecturer in the Department of Automatic Control and Systems Engineering at the University of Sheffield, Sheffield, UK. He is currently an Associate Professor in the School of Engineering at Westlake University, Hangzhou, China. His research interests lie in sensing, estimation, and control of robotic systems.

\end{IEEEbiography}

\end{document}